\documentclass{bmvc2k}

\usepackage{multirow}

\title{DAL: Dynamic Angular Loss for Imbalanced Medical Image Classification}

\addauthor{Salman Mohammad}{salman.mohammad@uzh.ch}{1}
\addauthor{Furkan Kasım}{furkan.kasim@uzh.ch}{1}
\addauthor{Manuel Günther}{siebenkopf@googlemail.com}{1}

\addinstitution{
 Department of Informatics\\
 University of Zurich\\
 Zurich, Switzerland
}

\runninghead{Mohammad et al}{DAL}

\def\eg{\emph{e.g}\bmvaOneDot}

\begin{document}

\maketitle

\begin{abstract}\noindent
Class imbalance remains a major obstacle in medical image classification, where rare but clinically important classes are often overshadowed by majority categories. Although current strategies for reducing bias, including angle-based softmax (ABS) losses, offer strong discriminative features, their behavior in settings with a small number of severely imbalanced classes remains insufficiently understood. Previous approaches that estimate the intrinsic parameters of ABS rely on the assumption of many classes, which does not hold in medical settings. We derive data-driven upper and lower bounds for the scale parameter and an upper bound for the angular margin using negative-angle statistics, and use these to drive a batch-adaptive angular margin (BAAM), which we combine with dynamically-weighted cross entropy (DWCE) to define our dynamic angular loss (DAL). Across four highly imbalanced medical image classification benchmarks spanning dermatology, ophthalmology, and gastrointestinal endoscopy, our method consistently achieves superior balanced accuracy and the lowest coefficient of variation of per-class recall. Although this entails a small
performance reduction on majority classes, the gains on minority categories are substantial, yielding a reliable and equitable overall classifier. Our source
code is available at: \url{https://github.com/msalman95/DAL}.
\end{abstract}

\section{Introduction}

Analysis of medical images is a central component of contemporary diagnostic workflows, supporting disease identification, early detection, and follow-up assessment across dermatology, ophthalmology, and other clinical domains. 
These tasks frequently require the interpretation of subtle and heterogeneous visual patterns, where diagnostic distinctions depend on fine structural and textural cues. Computer-aided diagnosis (CAD) assists medical practitioners to identify known diseases.
While some of these diseases appear regularly and many practitioners are able to identify them, the importance of CAD systems comes at highlighting rare cases that practitioners might not be familiar with.
Therefore, CAD systems should particularly focus on detecting rare cases, while still recognize common cases reasonably.

Many modern CAD systems are based on deep learning, due to its ability to learn task-specific representations directly from raw clinical images \citep{esteva2017dermatologist,yu2018melanoma}. 
Convolutional neural networks (CNNs) and related architectures have demonstrated strong performance on large-scale benchmarks, including dermatologist-level skin-cancer classification \citep{esteva2017dermatologist} and high-performing diabetic-retinopathy (DR) detection \citep{gulshan2016development}. However, disease distributions typically reflect population prevalence and clinical datasets are rarely balanced. 
Therefore, models operate under pronounced class disparities that influence learned representations \citep{johnson2019survey}.

Handling such class imbalance is a long-standing challenge in machine learning.
It is well-known that predictive models are biased toward majority classes, distort decision boundaries, and -- most severely for their practicability -- undermine minority-class recognition \citep{sun2009classification,krawczyk2016learning,rudd2016moon}. 
These issues become more pronounced in medical imaging, where minority categories often represent clinically meaningful but infrequent findings, \eg, melanomas occur notably less often than benign lesions, and exhibit substantial intra-class variability alongside high visual similarity to non-malignant categories \citep{gessert2019skin}. This often leads to reduced performance on minority classes, limiting the applicability of deep-learning systems in scenarios requiring reliable detection.

A variety of strategies for handling imbalanced data have been proposed, including class weighting, cost-sensitive learning, over- or undersampling, and data augmentation. 
Loss function modifications such as focal loss \citep{lin2017focalloss} and class-balanced loss \citep{cui2019balanced} reduce majority class dominance. \citet{buda2018systematic} showed that sampling techniques with CNN helped in improving overall performance under various degrees of imbalance.

Angular margin-based softmax formulations provide a mechanism to improve representation discriminability by imposing geometric constraints on the embedding space. 
Notably, ArcFace \citep{deng2022arcface} introduces an angular margin that encourages compact intra-class clusters and stronger inter-class separation.
Although these methods achieve strong results in large-scale benchmarks, they are generally developed under relatively balanced class distributions and rely on fixed global margin behavior. 
In highly imbalanced medical image settings, uniform angular constraints may be insufficient to maintain adequate separation for minority categories, allowing majority classes to dominate the learned embedding geometry. 

In this work, we analyze the behavior of angular-margin losses under class-imbalanced medical distributions and address this methodological gap. We derive data-driven upper and lower bounds for the scale parameter $s$ used in angular softmax formulations, providing principled guidance when the assumptions underlying prior adaptive scaling strategies \cite{zhang2019adacos} are violated. Additionally, we derive a theoretical upper bound for the margin parameter $m$. Based on this analysis, we introduce a dynamic angular loss (DAL) that integrates a batch-adaptive angular margin (BAAM) with dynamically weighted cross-entropy (DWCE) to improve robustness under long-tailed medical regimes. To assess performance equity, we additionally quantify variability across classes using the coefficient of variation (CoV). Experiments on ISIC \cite{codella2018skin,combalia2019bcn20000}, APTOS \cite{aptos2019-blindness-detection}, and Hyper-Kvasir~\cite{Borgli2020} demonstrate consistent improvements in balanced accuracy and minority-class sensitivity.  

\section{Related Work}

Research on class imbalance in medical image classification has motivated a variety of loss functions designed to mitigate majority-class dominance and improve minority-class sensitivity. A central line of work focuses on weighting-based objectives. 
Focal loss \citep{lin2017focalloss} suppresses the contribution of well-classified samples and places greater emphasis on harder cases, which are often those belonging to minority categories. 
More recently, class-wise difficulty-balanced loss \citep{sinha2020class} proposes adjusting sample contributions according to estimated class-wise difficulty, providing an adaptive mechanism that increases gradients for underperforming or underrepresented classes.
Complement cross-entropy \citep{kim2021imbalanced} similarly aims to stabilize optimization under skewed distributions by explicitly reducing confidence on incorrect classes rather than solely reinforcing the correct class. 

Another class of methods reformulates loss terms using estimates of effective class frequency. 
Class-balanced loss \citep{cui2019balanced} derives weighting coefficients from the effective number of samples, avoiding overly large weights in cases of severe imbalance. 
AutoBalance \citep{li2021autobalance} extends this principle by optimizing weighting parameters during training, demonstrating improvements without manual tuning. 
Label distribution-aware margin (LDAM) loss \citep{cao2019learning} proposes margin adjustments based on class frequencies, enforcing larger margins for minority classes to yield improved generalization under long-tailed distributions. 
Merge loss \citep{du2023merge} makes use of domain knowledge by aggregating predictions across closely related minority subclasses to stabilize optimization and enhance performance in extreme imbalance settings.
Other domain-specific loss formulations have also been introduced in dermatology and ophthalmology. 
For example, the adaptively weighted balance (AWB) loss \citep{yue2022toward} estimates class-wise weights dynamically during training and has shown gains for multi-center skin-lesion datasets with substantial imbalance. 
Similar developments have appeared in diabetic-retinopathy analysis, where angular-margin contrastive formulations have been used to improve discriminative separation between DR grades \citep{zhu2023supervised}.
In parallel, several data-centric imbalance strategies have been developed. 
Balanced MixUp \citep{galdran2021balanced} adapts MixUp to produce synthetic samples that preserve the structure of minority classes in medical image settings. 
\citet{scholz2024imbalance} provide a systematic study demonstrating that imbalance-sensitive loss functions consistently outperform naive cross-entropy across a range of medical classification tasks, reinforcing the need for specialized treatments of clinical class skew.

Beyond weighting and frequency-based corrections, angular margin-based losses provide an orthogonal mechanism by enforcing geometric constraints on the embedding space. 
Most such methods are developed for the face recognition task \cite{wang2021survey}, where abundances of classes with large amounts of samples are available for training \cite{zhu2021webface260m}.
ArcFace \citep{deng2022arcface} adds a fixed additive angular margin to encourage compact intra-class clusters and increased inter-class separation. 
AdaCos \citep{zhang2019adacos} removes the need for manually tuned scale parameters by introducing adaptive scaling of cosine logits. 
AdaFace \citep{kim2022adaface} further incorporates quality-dependent margins that adjust the angular decision boundary based on sample reliability. 
Such methods are increasingly explored in medical domain. 
For example, \citet{alirezazadeh2023boosted} applied them to histopathology-based breast cancer classification, and \citet{prakash2025advancing} demonstrate their potential for improving discriminative medical image representations. 
However, existing angular-margin approaches assume balanced class distributions, leaving a methodological gap that we address in our work.
\section{Approach}
We extend angular-margin based classification by addressing the limitations of ArcFace \cite{deng2022arcface} under severe class imbalance. 
We provide a data-driven generalization of the AdaCos \cite{zhang2019adacos} scale parameter $s$ for stabilizing \emph{angle-based softmax} (ABS), and batch-level mechanisms that compensate for imbalance through dynamic sample weighting or adaptive angular margins. We base our analysis on the angle statistics of the ISIC 2018 \cite{codella2019skin} training set, and observe similar distributions across other datasets, so the same approach applies to all of them.

\subsection{Generalized Scale Parameter}\label{sec:scale}

\begin{figure}[h]\centering
\includegraphics[width=0.75\linewidth]{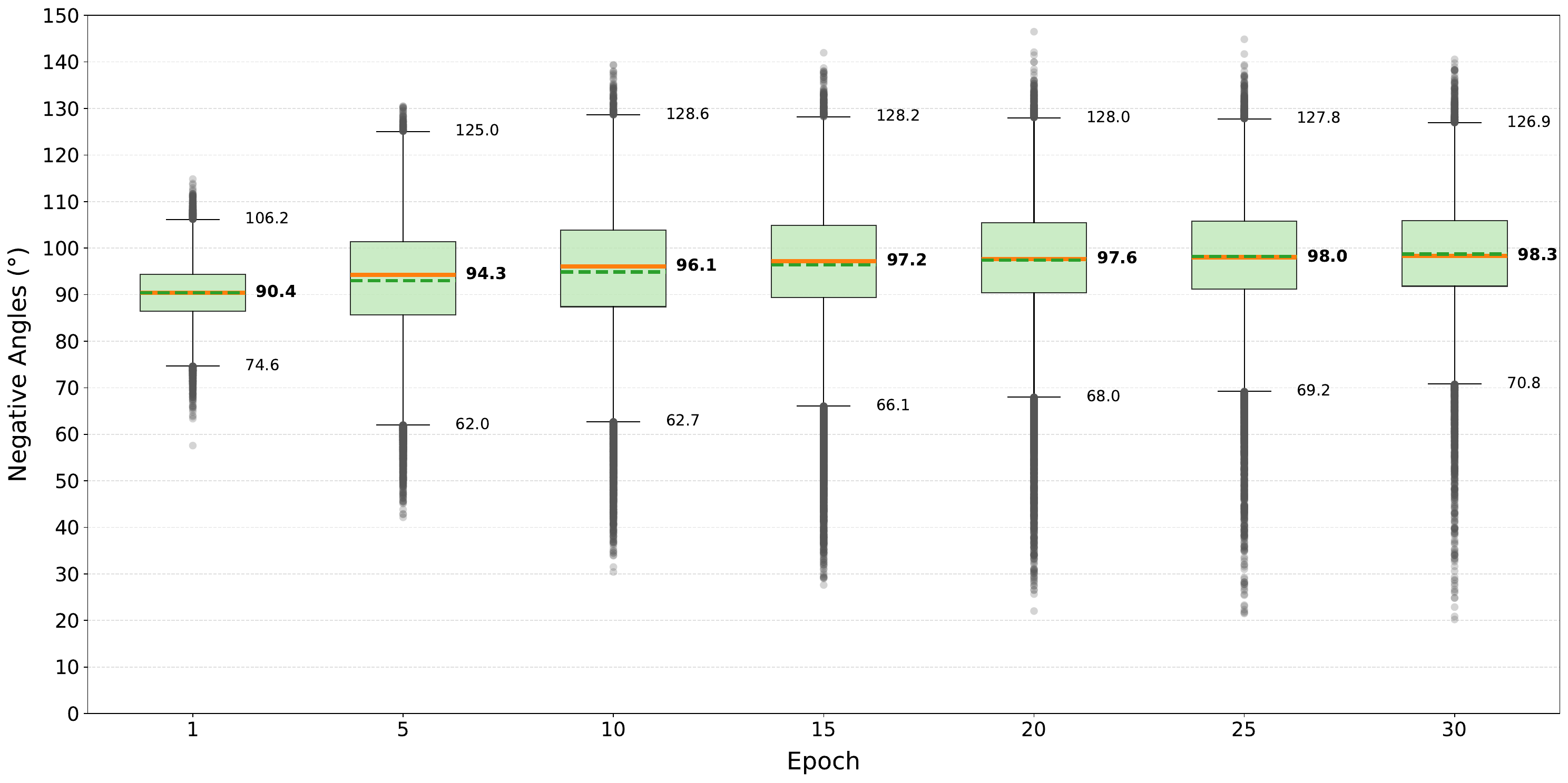}
\Caption[fig:negative-angles-s4]{Negative Angles per Epoch}{
This figure shows box plots of the distribution of ArcFace angles $\theta_{i,j}$ for negative classes on the ISIC 2018 training set during training.
}
\end{figure}

Let $\mathbf{x}_i$ denote the embedding of sample $i$ with label $1\leq y_i\leq C$, and $\theta_{i,j}$ be the angle between $\mathbf{x}_i$ and classifier weight $\mathbf{W}_j$. 
The \emph{angle-based softmax} (ABS) \cite{deng2022arcface} is defined as:
\begin{equation}
\label{eq:adacos}
    P_{i,y_i}
    = \frac{\exp\bigl(s\cos(\theta_{i,y_i}+m)\bigr)}
           {\exp\bigl(s\cos(\theta_{i,y_i}+m)\bigr) + B_i},
    \qquad
    B_i = \sum_{j\ne y_i} \exp\bigl(s\cos\theta_{i,j}\bigr),
\end{equation}
where $B_i$ represents angles $\theta_{i,j}$ of all $C-1$ negative classes $j\neq y_i$.
Assuming a margin of $m=0$, AdaCos \cite{zhang2019adacos} determines the optimal scale $s$ by maximizing the curvature of $P_{i,y_i}(\theta)$ at a chosen central angle $\theta_0$, leading to:
\begin{equation}
    \label{eq:adacos-s}
    s_0 = \frac{\log B_i}{\cos\theta_0}.
\end{equation}
This original AdaCos formulation assumes that negative class angles $\theta_{i,j}$ are concentrated at $90\degree$, implying $\forall j\neq y_i: \cos\theta_{i,j}\approx 0$ and yielding $B_i \approx C-1$ \cite{zhang2019adacos} along with the closed-form scale $s_{\mathrm{AdaCos}} = \sqrt{2}\log(C-1)$. 
While such assumptions are reasonable for face recognition, in our setting with fewer classes and smaller amounts of samples, however, negative angles span a substantially wider interval (see \figref{fig:negative-angles-s4}), which violates the assumption $\theta_{i,j}\approx90\degree$.\footnote{Distributions for other datasets are given in the supplementary material.}

To obtain a more faithful scale estimate, we first assume that all negative angles have a fixed value $\theta_{i,j}$, which we later relax to provide upper and lower bounds for $s$ derived from batch statistics.
We rewrite the negative term in \eqref{eq:adacos}: 
\begin{equation}
    \label{eq:s-neg}
    B_i \approx (C-1)\exp(s\cos\theta_{i,j}).
\end{equation}
Substituting this approximation into \eqref{eq:adacos-s} yields the generalized closed-form scale. 
Specifically, taking the logarithm of \eqref{eq:s-neg} and replacing $\log B_i$ in \eqref{eq:adacos-s} results in: 
\begin{align}
    \label{eq:logBi}
    \log B_i & \approx \log(C-1) + s_0 \cos\theta_{i,j}\,,
&   s_0 &= \frac{\log(C-1) + s_0 \cos\theta_{i,j}}{\cos\theta_0}.
\end{align}
%
Solving for $s_0$ leads to the generalized closed-form scale:
\begin{equation}
    s_0
    = \frac{\log(C-1)}
           {\cos\theta_0 - \cos\theta_{i,j}}
\end{equation}
Following the original AdaCos formulation \cite{zhang2019adacos}, we set the reference positive angle to $\theta_0=45\degree$, the geometric midpoint of the feasible positive-angle range $[0,90\degree]$. 
When using $C=7$ and the empirical negative-angle interval observed in \figref{fig:negative-angles-s4}, namely $\theta_{i,j}\in[65\degree,125\degree]$, the above expression produces a data-driven scale range of $s_0 \in [1.40,\, 6.30]$.
Based on this interval, we can select the mid-range value $s=4$, which provides stable optimization without compromising discriminability.  For other datasets, similar data distribution can be observed.

\begin{figure}[t]\centering
\includegraphics[width=0.99\linewidth]{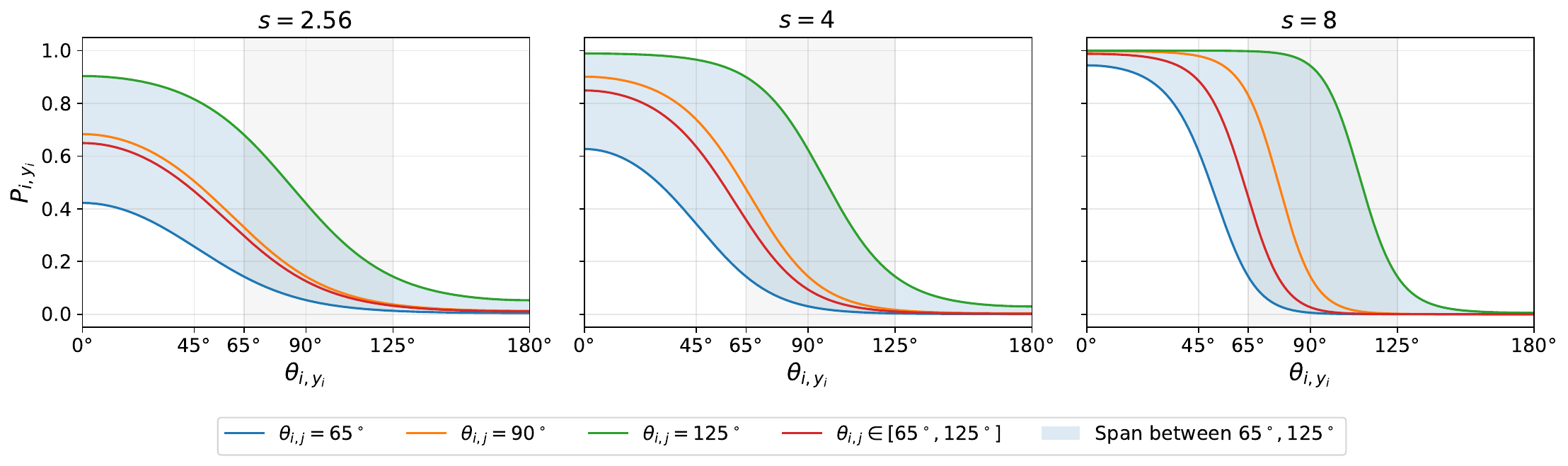}
\Caption[fig:arcface-s-comparison]{Negative Angle Placement}{
    We show the effect of negative angles $\theta_{i,j}; j\neq y_i$ on the positive-class probability $P_{i,y_i}$ across different scale values $s$ for $C=7$.
    The curves correspond to four negative-angle configurations: three fixed cases where $\forall j\neq y_i:\,\theta_{i,j}=65^\circ$, $90^\circ$, or $125^\circ$, and the most realistic distributed case in which $\theta_{i,j}$ is linearly assigned within the interval $[65^\circ,125^\circ]$ across negative classes.
}
\end{figure}

\figref{fig:arcface-s-comparison} illustrates how the positive-class probability $P_{i,y_i}$ evolves as a function of the positive angle $\theta_{i,y_i}$ for different choices of the scale parameter $s$ and several representative negative angle configurations. 
When $s$ is chosen too small, such as $s \approx 2.56$ for $C=7$ classes using the original AdaCos formulation in \eqref{eq:adacos-s}, the probability remains significantly below~1 even when the embedding is perfectly aligned with its class center, i.e.\ $\theta_{i,y_i} \approx 0$. 
This occurs because $s\cos\theta_{i,y_i}$ in \eqref{eq:adacos} cannot sufficiently dominate the negative logits, even when negative angles $\theta_{i,j}$ are large. 
Conversely, when $s$ is excessively large, the probability saturates too early: even for relatively large positive angles (far from the class center $\mathbf W_{y_i}$), the model assigns very high confidence whenever negative angles are moderately high. 
Such behavior undesirably compresses the angular decision margin and leads to overconfident predictions and poorly calibrated gradients.
Within the empirically observed negative-angle interval $[65^\circ,125^\circ]$ (see \figref{fig:negative-angles-s4}), the choice
$s=4$ provides a favorable balance. 
As highlighted by the shaded region in \figref{fig:arcface-s-comparison}, this scale yields well-behaved probability curves: the model achieves high confidence only when $\theta_{i,y_i}$ is sufficiently small, while avoiding premature probability saturation for larger angles. This intermediate scaling therefore produces stable optimization dynamics and preserves angular discriminability across classes.

\subsection{Batch-Adaptive Angular Margin and Dynamic Weighting}

Under class imbalance, ABS suffers from two limitations. 
First, majority classes dominate the optimization because they appear more frequently in the loss and second, a fixed angular margin $m$ imposes identical geometric constraints on all classes, irrespective of their representation in the data.
To address both issues simultaneously, we combine a \emph{batch-adaptive angular margin} (BAAM) with \emph{dynamic weighted cross-entropy} (DWCE) loss.

\subsubsection{Admissible Margin Range}\label{sec:Margin}
The scale parameter $s$ derived in \secref{sec:scale} is anchored at the
reference angle $\theta_0 = 45\degree$, the midpoint of the feasible positive-angle
range $[0, 90\degree]$. The scale is chosen under $m = 0$ such that a sample at
$\theta_0$ lies just inside the high-confidence region of the probability
curve, i.e., the positive logit $s\cos\theta_0$ dominates the negative-class
log-sum. When a non-zero margin~$m$ is added, the effective positive angle of
that sample shifts from $\theta_0$ to $\theta_0 + m$ in the softmax formula \eqref{eq:adacos}.
We require this shifted effective angle to remain within the high-confidence
region, so that the assumption under which $s$ was selected is preserved.

The boundary of the high-confidence region is the angle at which the positive
and negative contributions to the softmax denominator balance, i.e., where
$P_{i,y_i} = \tfrac12$. Below this transition angle $\theta_{trans}$
(for $\theta < \theta_{trans}$, where the sample sits closer to its class
center) the positive logit dominates and the sample is classified with
confidence; above it, the negative-class log-sum $B_i$ in \eqref{eq:adacos}
dominates and confidence falls below chance. Setting the positive logit equal
to the log of the negative-class contribution:
\begin{equation}
    \label{eq:transition_balance}
    s\cos\theta_{trans} \;=\; \ln(C - 1) + s\cos\theta_{i,j}\,,
\end{equation}
and solving for the transition angle gives:
\begin{equation}
    \label{eq:theta_trans}
    \theta_{trans}
    \;=\;
    \arccos\!\left( \cos\theta_{i,j} + \frac{\ln(C - 1)}{s} \right).
\end{equation}
The consistency requirement $\theta_0 + m \leq \theta_{trans}$
then yields the admissible margin bound
\begin{equation}
    \label{eq:m_max_bound}
    m \;\leq\; \theta_{trans} - \theta_0 \;=:\; m_{\max}\,.
\end{equation}

Because $\theta_{trans}$ increases monotonically with the negative angle
$\theta_{i,j}$, the ceiling $m_{\max}$ grows as negatives become better
separated. Although $\theta_{i,j}$ can exceed $90\degree$ in practice, we clamp
it at the orthogonality threshold $\theta_{i,j}=90\degree$
($\cos\theta_{i,j}=0$): beyond this point the sample is already well separated
from the negative center, and we do not credit the additional separation. For
$C = 7$ and our derived $s = 4$, this gives $\theta_{trans} \approx 63.4\degree$
and the upper endpoint $m_{\max} \approx 18.4\degree \approx 0.32$\,rad.

At the opposite, worst-case end---the most confusable negatives in a batch,
consistent with the extreme used for the scale derivation in
\secref{sec:scale}---$\theta_{trans}$ drops below $\theta_0$, so $m_{\max} < 0$
and no positive margin preserves the confidence assumption. We therefore set the
lower endpoint to $m_{\min} = 0$, the only value admissible across all batch
configurations. For $C = 7$ and $s = 4$, the admissible margin range is thus
$[m_{\min},\, m_{\max}] = [0,\,0.32]$\,rad.

\subsubsection{Batch-Adaptive Angular Margin (BAAM)}
Let a training batch of size $N$ contain $N_c$ samples of class
$1 \leq c \leq C$. To assign per-class margins within the admissible range
$[0, m_{\max}]$ derived above, we use a simple inverse-frequency mapping
that assigns larger margins to under-represented classes. We compute a raw
inverse-frequency score $t_c^{\mathrm{raw}}$, normalize it batch-wise, and
map the normalized score to the angular margin range:
\begin{align}
    \label{eq:BAAM}
    t_c^{\mathrm{raw}} &= \log\!\left(1 + \frac{1}{N_c}\right), &
    m_c &= m_{\min} + (m_{\max} - m_{\min})\, \frac{t_c^{\mathrm{raw}} - t_{\min}}{t_{\max} - t_{\min}}\,.
\end{align}
Each sample inherits the margin of its ground-truth class, replacing the
fixed margin $m$ in \eqref{eq:adacos} by $m_{y_i}$. Larger $m_{y_i}$ values
compel minority classes to form more compact discriminative embeddings,
mitigating geometric dominance by majority classes. We adopt the
$\log(1 + 1/N_c)$ form as a simple monotonic transformation of inverse
class frequency: it is smooth, bounded for finite $N_c$, and maps the head
class ($N_c$ large) close to zero while preserving discrimination among
rare classes ($N_c$ small). We use $[m_{\min}, m_{\max}] = [0, 0.2]$\,rad in
all our experiments, well within the admissible bound
$m_{\max} \approx 0.32$\,rad given by \eqref{eq:m_max_bound}. The most
frequent class in each batch receives $m_c = m_{\min} = 0$, while the rarest
class receives $m_c = m_{\max}$; intermediate classes are placed in between
by the batch-normalized mapping. At inference, the ground-truth class is
unknown and we set $m = 0$.

\subsubsection{Dynamic Weighted Cross-Entropy (DWCE)}
While adaptive margins modify the representation geometry, the optimization process itself remains sensitive to class frequencies.
To ensure balanced gradient flow, we incorporate a dynamic weighting factor computed from the batch composition.
For each batch, we follow standard class weighting techniques \cite{palechor2023protocols} and define the class weights $w_c$ so that minority classes receive proportionally larger loss contributions.\footnote{When the number of samples per class in a batch is $N_c=0$, we assume $\omega_c=0$.}
The dynamic weighted cross-entropy (DWCE) loss is:
\begin{align}
    \label{eq:DWCE}
    \mathcal{L}_{\mathrm{DWCE}} &= -\frac{1}{N}\sum_i \omega_{y_i}\log P_{i,y_i}\,, &
    \omega_c&=\frac{N}{C\,N_c}\,.
\end{align}
Finally, the proposed \emph{dynamic angular loss} (DAL) uses ABS to compute $P_{i,y_i}$ via \eqref{eq:adacos} with our derived $s=4$ and BAAM, and minimizes $\mathcal{L}_{\mathrm{DWCE}}$ from \eqref{eq:DWCE}.

\section{Experiment}

\subsection{Datasets and Evaluation}

\begin{figure*}[t]
{
\subfigure[ISIC 2018]{\includegraphics[ width=.20\textwidth]{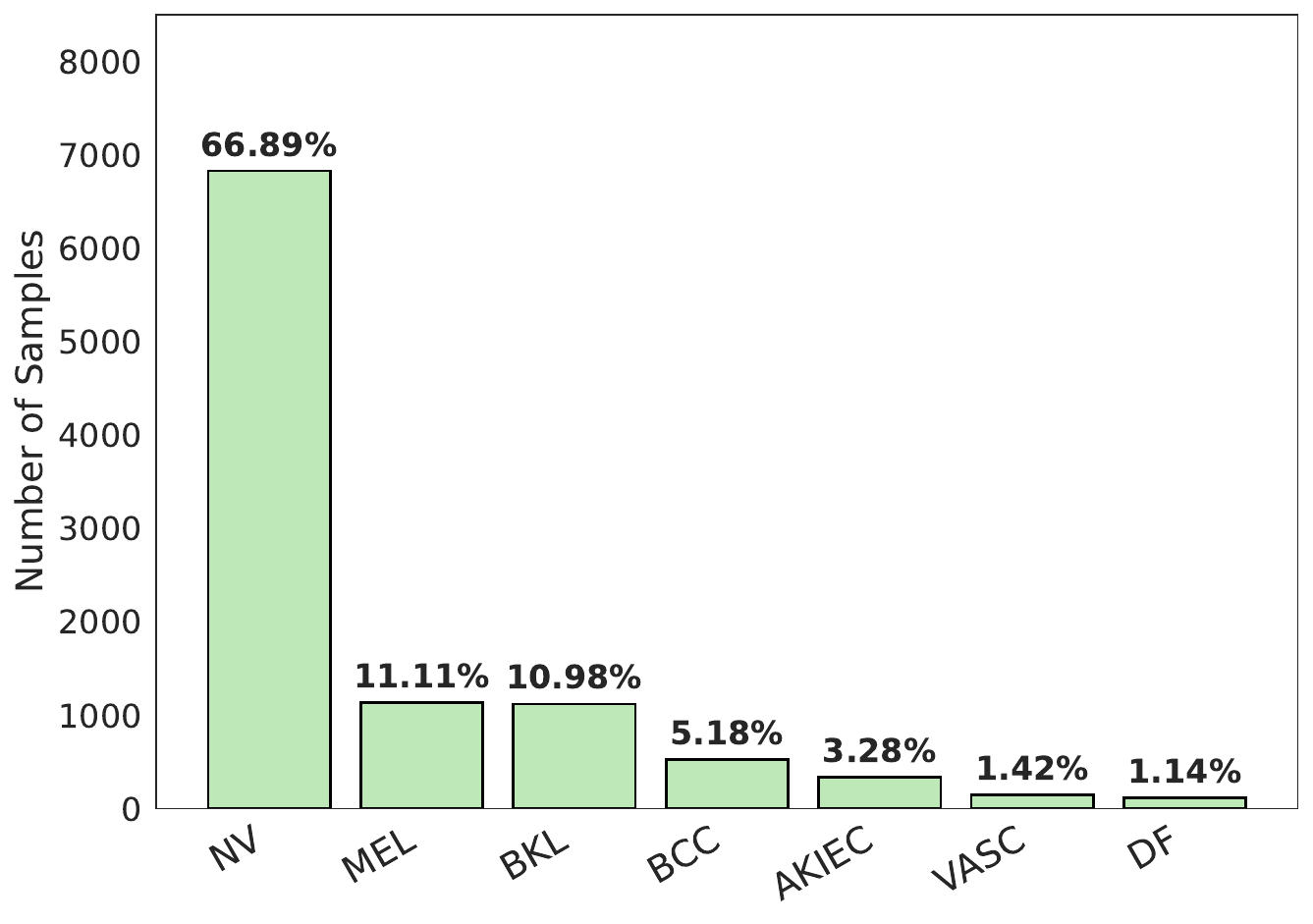}}
\subfigure[ISIC 2019]{\includegraphics[width=.20\textwidth]{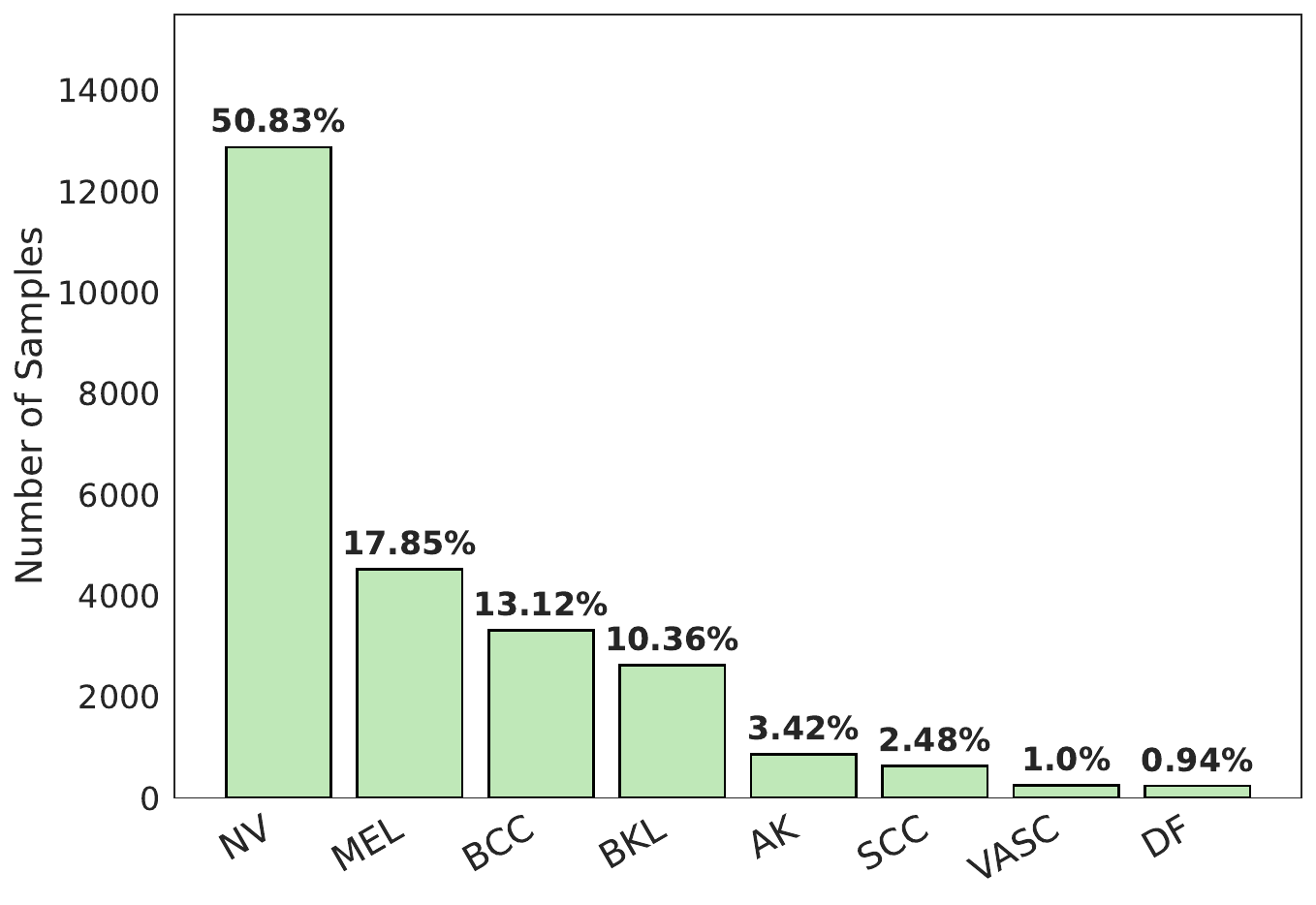}}
\subfigure[APTOS]{\includegraphics[width=.20\textwidth]{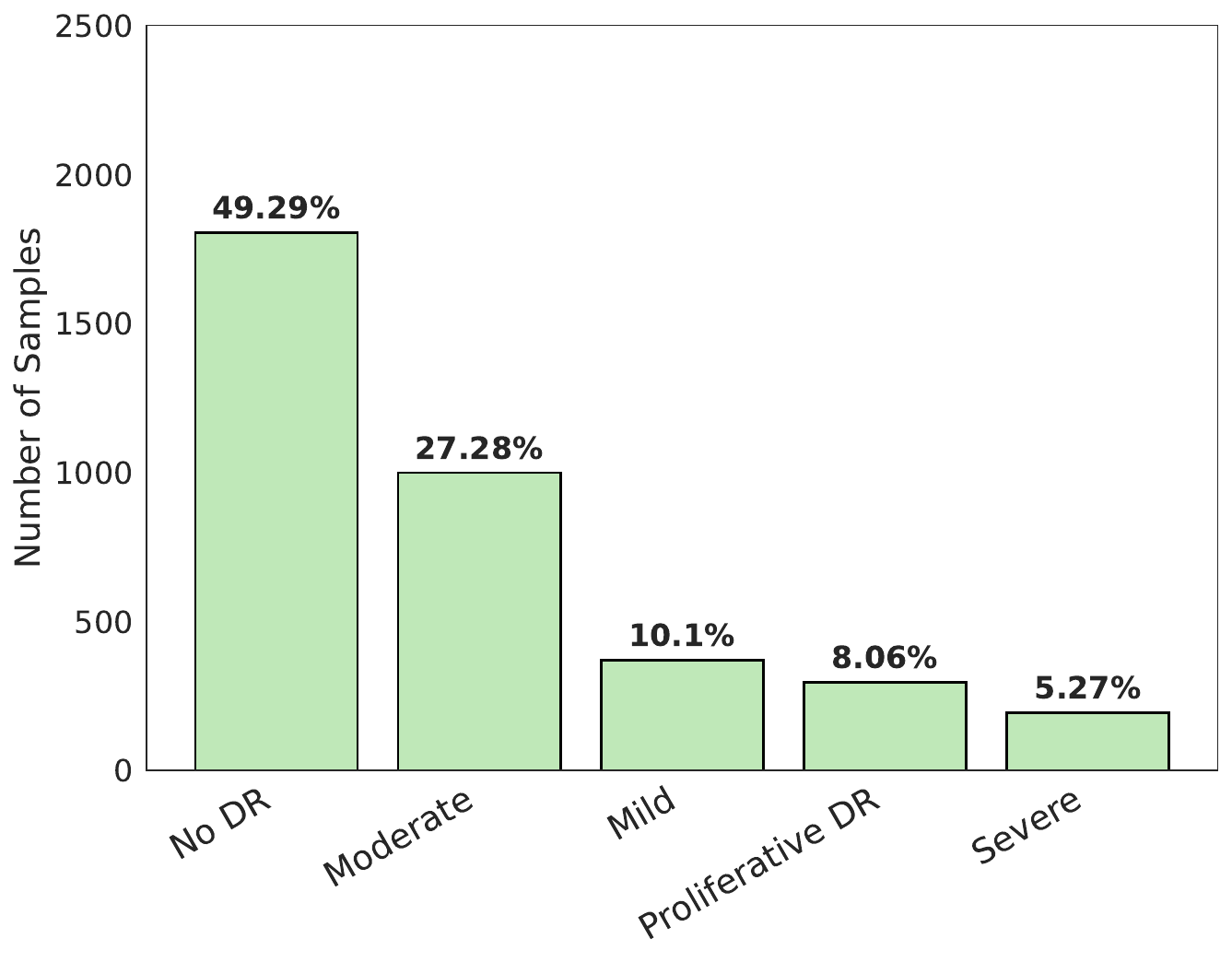}}
\subfigure[Hyper-kvasir]{\includegraphics[width=.30\textwidth]{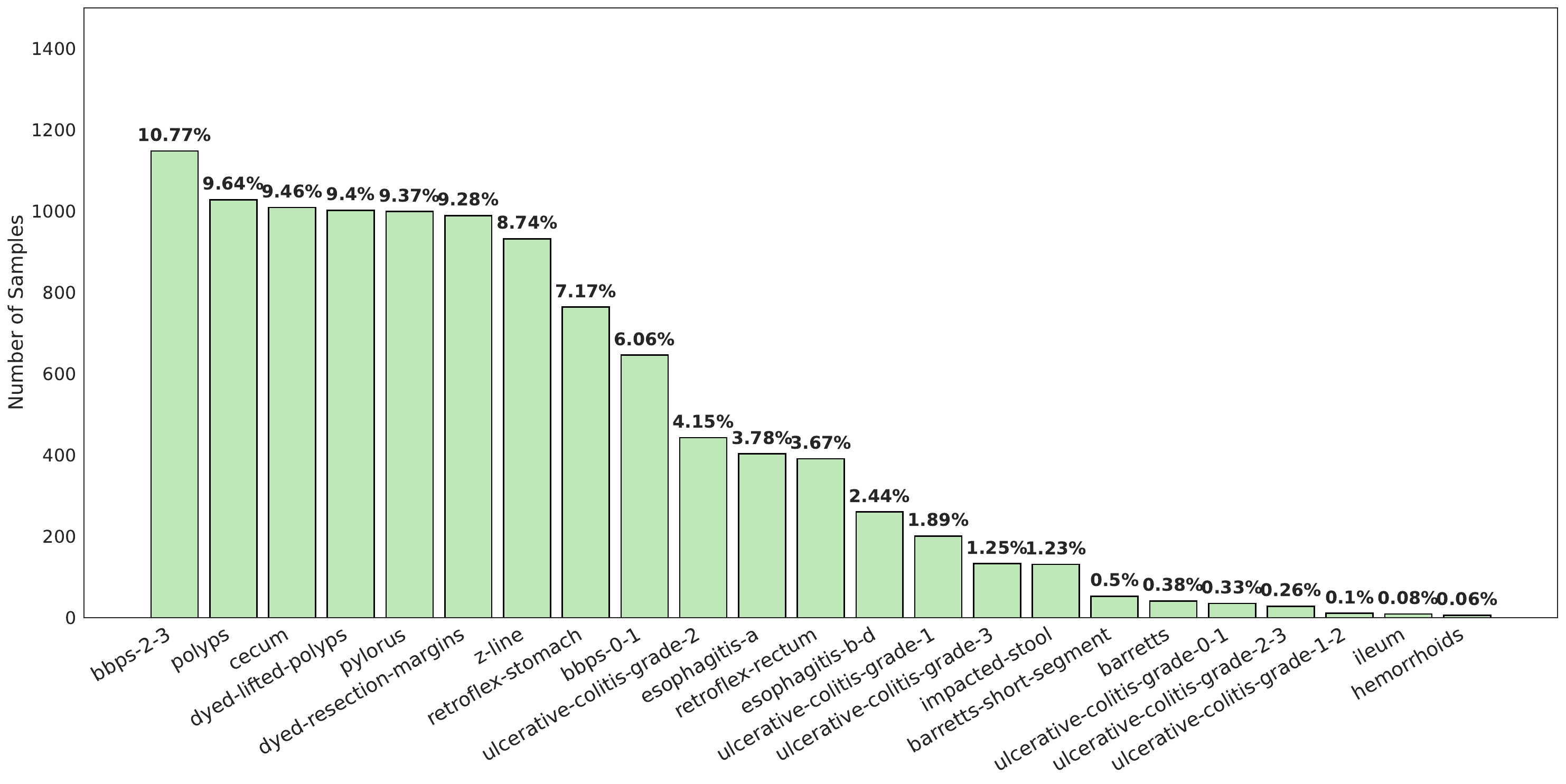}}
}

{\Caption[fig:dataset-distributions]{Class Distribution}{
    For each dataset, classes are ordered in decreasing frequency and annotated with their relative proportion (in \%) above each bar.
}}

\end{figure*}

We conduct experiments on four publicly available medical image classification datasets:
ISIC 2018~\cite{tschandl2018ham10000, codella2019skin}, ISIC 2019~\cite{tschandl2018ham10000,codella2018skin,combalia2019bcn20000} for skin lesion classification, APTOS 2019~\cite{aptos2019-blindness-detection} for diabetic retinopathy grading and Hyper-Kvasir~\cite{Borgli2020} for Gastrointestinal findings. All four datasets exhibit high imbalance distribution. For ISIC 2018 and ISIC 2019, we use the official subject-independent test set for final evaluation.
For APTOS 2019 and Hyper-Kvasir datasets, no official test set is provided, so we perform an 80/20 stratified train/test split.
In all cases, we divide the training portion into 80\% training and 20\% validation data for model selection using stratified sampling. 
We report accuracy (Acc), balanced accuracy (BA), macro-averaged F1-score, and the Index of Balanced Accuracy (IBA$_\alpha$)~\cite{garcia2009index} for $\alpha =1$.\footnote{We also evaluated IBA$_\alpha$ with other $\alpha$  values, and conclusions are identical.} 
Additionally, we use the coefficient of variation (CoV) of per-class recall~\cite{formby1999coefficient,chen2023guide} to quantify performance stability across classes. 
All experiments are repeated five times with different seeds, and the mean is reported. For Hyper-Kvasir, we use $s = 6$, based on the admissible scale range of $[2.4, 10.9]$, and report results over head, medium, and tail splits defined by \citet{Borgli2020}.

\subsubsection{Implementation Details}
We took the original implementations of various competitors and applied an identical training strategy.
All models are implemented in PyTorch~\cite{paszke2019pytorch} using a ResNet-18 backbone~\cite{he2016deep} pre-trained on ImageNet~\cite{russakovsky2015imagenet}. 
Our images are resized to $224 \times 224$. 
Following~\cite{deng2022arcface}, we replace the original fully connected layer with a lightweight head consisting of dropout, a linear projection from the 512-dimensional backbone output to a 128-dimensional embedding, batch normalization, and a classification layer $\mathbf W$. 
Training is performed using SGD with a learning rate of 0.001, momentum of 0.9, and weight decay of 0.0001 for 30 epochs with a batch size of $N=128$. 
The backbone is frozen for the first five epochs and then fine-tuned jointly with the classification head. 
We select the epoch achieving the highest balanced accuracy on the validation set.

\begin{table}[bt]
\begin{center}
\scriptsize

\begin{tabular}{@{}l||c|c|c|c|c||c|c|c|c|c@{}}
\multirow{2}{*}{\bfseries Method}
& \multicolumn{5}{c||}{\bf ISIC 2018} & \multicolumn{5}{c}{\bf ISIC 2019} \\
\cline{2-11}
 & \bfseries Acc$\uparrow$
 & \bfseries F1$\uparrow$
 & \bfseries BA$\uparrow$
 & \bfseries IBA$_1$$\uparrow$
 & \bfseries CoV$\downarrow$
 & \bfseries Acc$\uparrow$
 & \bfseries F1$\uparrow$
 & \bfseries BA$\uparrow$
 & \bfseries IBA$_1$$\uparrow$
 & \bfseries CoV$\downarrow$ \\
\hline
Cross-Entropy
 & 0.7779 & 0.6182 & 0.6015 & 0.3887 & 0.2612 
 & 0.6378 & 0.4633 & 0.4562 & 0.2551 & 0.4281 \\
AdaCos \cite{zhang2019adacos}
 & 0.7741 & 0.5899 & 0.5897 & 0.4097 & 0.2910     
 & 0.6388 & 0.3744 & 0.3827 & 0.2243 & 0.7255 \\
ArcFace \cite{deng2022arcface}
 & 0.7857 & 0.6375 & 0.6407 & 0.4187 & 0.2101
 & 0.6434 & 0.4517 & 0.4239 & 0.2424 & 0.5614 \\
Focal Loss \cite{lin2017focalloss}
 & 0.7722 & 0.6195 & 0.6095 & 0.3969 & 0.2473
 & 0.6466 & 0.4792 & 0.4687 & 0.2634 & 0.4193 \\
CCE Loss \cite{kim2021imbalanced}
 & 0.7738 & 0.6162 & 0.6084 & 0.3991 & 0.2339
 & 0.6467 & 0.4756 & 0.4576 & 0.2574 & 0.4377 \\
LDAM Loss \cite{cao2019learning}
 & \textbf{0.7864} & 0.6603 & 0.6678 & 0.4547 & 0.2043
 & 0.6319 & 0.4759 & 0.4798 & 0.2869 & 0.3390 \\
\hline
CB Loss \cite{cui2019balanced}
 & 0.7763 & \textbf{0.6680} & 0.6730 & 0.4659 & 0.1300 
 & \bf 0.6472 & 0.5105 & 0.4970 & 0.2881 & 0.3548 \\
PF Loss \cite{du2021parameter}
 & 0.7357 & 0.6465 & 0.6941 & 0.4936 & \textbf{0.0930}
 & 0.6308 & 0.5029 & 0.5233 & 0.3040 & 0.2533 \\
AWB Loss \cite{yue2022toward}
 & 0.7312 & 0.6368 & 0.6500 & 0.4375 & 0.1369
 & 0.6292 & \bf 0.5134 & 0.5045 & 0.2937 & 0.3439 \\
DCAL \cite{kim2025long}
 & 0.7689 & 0.6313 & 0.6363 & 0.4274 & 0.2054
 & 0.6328 & 0.4886 & 0.4812 & 0.2711 & 0.3463 \\
DWCE Loss \eqref{eq:DWCE}
 & 0.7630 & 0.6606 & 0.6790 & 0.4718 & 0.1340      
 & 0.6392 & 0.5089 & 0.5029 & 0.2891 & 0.3232 \\
\hline
DAL (Ours)
 & 0.7443 & 0.6490 & \bf 0.7348 & \bf 0.5470 & 0.0972
 & 0.6214 & 0.5063 & \bf 0.5545 & \bf 0.3343 & \bf 0.1877 \\

\end{tabular}

\vspace{1.5ex}

\begin{tabular}{@{}l||c|c|c|c|c||c|c|c|c|c@{}}
\multirow{2}{*}{\bfseries Method}
& \multicolumn{5}{c||}{\bf APTOS} & \multicolumn{5}{c}{\bf Hyper-Kvasir} \\
\cline{2-11}
 & \bfseries Acc$\uparrow$
 & \bfseries F1$\uparrow$
 & \bfseries BA$\uparrow$
 & \bfseries IBA$_1$$\uparrow$
 & \bfseries CoV$\downarrow$
 & \bfseries Acc$\uparrow$
 & \bfseries F1$\uparrow$
 & \bfseries BA$\uparrow$
 & \bfseries IBA$_1$$\uparrow$
 & \bfseries CoV$\downarrow$ \\
\hline\hline
Cross-Entropy
 & 0.8175 & 0.6163 & 0.6047 & 0.4602 & 0.4842
 & \bf 0.8931 & 0.5751 & 0.5777 & 0.5110 & 0.7275 \\
AdaCos \cite{zhang2019adacos}
 & 0.7722 & 0.5246 & 0.5291 & 0.3994 & 0.6138
 & 0.8810 & 0.5713 & 0.5691 & 0.5004 & 0.7211 \\
ArcFace \cite{deng2022arcface}
 & 0.8003 & 0.6262 & 0.6136 & 0.4393 & 0.3789
 & 0.8971 & 0.5940 & 0.5910 & 0.5101 & 0.6788 \\
Focal Loss \cite{lin2017focalloss}
 & \bf 0.8292 & 0.6286 & 0.6111 & 0.4664 & 0.4847
 & 0.8926 & 0.5752 & 0.5803 & 0.5123 & 0.7204 \\
CCE Loss \cite{kim2021imbalanced}
 & 0.7999 & 0.5794 & 0.5748 & 0.4309 & 0.5335
 & 0.8831 & 0.5791 & 0.5812 & 0.5057 & 0.6912 \\
LDAM Loss \cite{cao2019learning}
 & 0.8002 & 0.6321 & 0.6287 & 0.4697 & 0.3538
 & 0.8834 & 0.5813 & 0.5845 & 0.5091 & 0.6849 \\
\hline
CB Loss \cite{cui2019balanced}
 & 0.8028 & 0.6319 & 0.6439 & 0.4621 & 0.3064
 & 0.8870 & 0.5813 & 0.5899 & 0.5108 & 0.6821 \\
PF Loss \cite{du2021parameter}
 & 0.7902 & 0.6349 & 0.6410 & 0.4653 & 0.3234
 & 0.8790 & 0.5841 & 0.5948 & 0.5122 & 0.6613 \\
AWB Loss \cite{yue2022toward}
 & 0.8016 & \bf 0.6491 & 0.6481 & 0.4674 & 0.2989
 & 0.8836 & 0.5854 & 0.5926 & 0.5123 & 0.6911 \\
DCAL \cite{kim2025long}
 & 0.8018 & 0.6224 & 0.6118 & 0.4405 & 0.3879
 & 0.8886 & 0.5783 & 0.5840 & 0.5096 & 0.7013 \\
DWCE Loss \eqref{eq:DWCE}
 & 0.8122 & 0.6484 & 0.6384 & 0.4656 & 0.3504
 & 0.8816 & 0.5875 & 0.5914 & 0.5077 & 0.6799 \\
\hline
DAL (Ours)
 & 0.7993 & 0.6467 & \bf 0.6756 & \bf 0.4923 & \bf 0.2381
 & 0.8880 & \bf 0.6049 & \bf 0.6125 & \bf 0.5312 & \bf 0.6412 \\

\end{tabular}

\end{center}

\Caption{Test set Performance Across Four Imbalanced Benchmarks}{Methods are separated into non- or implicitly-weighted losses, explicitly class-weighted losses, and Ours. Mean over 5 runs. Best results per dataset and metric are highlighted in \textbf{bold}.}
\label{tab:test-results}
\end{table}
\subsection{Results}
\subsubsection{Overall Performance}
\tabref{tab:test-results} summarizes the aggregated results.\footnote{The same table with uncertainty estimates (mean $\pm$ std) is provided in the supplementary material.} DAL achieves the
\textbf{highest balanced accuracy} on all four datasets, improving over the next-best
competitor by between 1.8\% on Hyper-Kvasir and 4.1\% on ISIC 2018. The same
trend holds for IBA$_1$, where DAL obtains the best value on every dataset. DAL
also yields the lowest CoV of per-class recall on three of the four datasets,
and is a close second on ISIC 2018, reaching 0.097 against 0.093 for PF, which
indicates the most uniform behavior across classes. Macro F1 is instead led by
AWB on ISIC 2019 and APTOS and by CB on ISIC 2018, while DAL attains the best F1
on Hyper-Kvasir. Finally, although the explicitly class-weighted losses in the
lower part of \tabref{tab:test-results} already improve balanced accuracy and
CoV, DAL provides further gains over the DWCE baseline on both metrics across
all datasets.

\begin{table}[p]
\begin{center}
\scriptsize

\subtable[ISIC 2018\label{tab:perclass-isic-2018}]{
  \begin{tabular}{l||c|c|c|c|c|c|c}
  \bfseries Method
  & \bfseries NV
  & \bfseries MEL
  & \bfseries BKL
  & \bfseries BCC
  & \bfseries AK
  & \bfseries VASC
  & \bfseries DF \\
  \it Class Frequency
  & \it 66.89\,\%
  & \it 11.11\,\%
  & \it 10.98\,\%
  & \it 5.18\,\%
  & \it 3.28\,\%
  & \it 1.42\,\%
  & \it 1.14\,\% \\
  \hline\hline
  Cross-Entropy
  & \textbf{0.9106} & 0.5438 & 0.6497 & 0.5462 & 0.6744 & 0.4988 & 0.3868 \\
  AdaCos \cite{zhang2019adacos}
  & 0.9011 & 0.6071 & 0.6240 & 0.6009 & 0.6452 & 0.3897 & 0.3458 \\
  ArcFace \cite{deng2022arcface}
  & 0.9058 & 0.5492 & 0.6381 & 0.6091 & 0.6891 & 0.5748 & 0.5080 \\
  Focal Loss \cite{lin2017focalloss}
  & 0.9047 & 0.5380 & 0.6230 & 0.5440 & 0.7162 & 0.5001 & 0.4404 \\
  CCE Loss \cite{kim2021imbalanced}
  & 0.9065 & 0.5166 & 0.6421 & 0.5627 & 0.7132 & 0.5619 & 0.3561 \\
  LDAM Loss \cite{cao2019learning}
  & 0.9073 & 0.5480 & 0.5569 & 0.6451 & 0.7209 & 0.7128 & 0.5227 \\
  \hline
  CB Loss \cite{cui2019balanced}
  & 0.8649 & 0.6562 & 0.6775 & 0.6150 & 0.6558 & 0.6285 & 0.6591 \\
  PF Loss \cite{du2021parameter}
  & 0.7863 & 0.6561 & 0.6368 & 0.6496 & \textbf{0.7581} & 0.6857 & 0.6863 \\
  AWB Loss \cite{yue2022toward}
  & 0.7924 & \textbf{0.6912} & \textbf{0.6820} & \textbf{0.6557} & 0.5548 & 0.6176 & 0.5817 \\
  DCAL \cite{kim2025long}
  & 0.8229 & 0.6832 & 0.6744 & 0.6021 & 0.6314 & 0.6010 & 0.4551 \\
  DWCE Loss \eqref{eq:DWCE}
  & 0.8646 & 0.6514 & 0.6671 & 0.6194 & 0.6593 & 0.6314 & 0.6698\\
  \hline
  \bf DAL (Ours)
  & 0.7780 & \textbf{0.6912} & 0.6737 & 0.6344 & 0.7163 & \textbf{0.8171} & \textbf{0.8091} \\[-2ex]
  \end{tabular}}

\vspace{1.5ex}

\subtable[ISIC 2019\label{tab:perclass-isic-2019}]{
  \begin{tabular}{l||c|c|c|c|c|c|c|c}
  \bfseries Method
  & \bfseries NV
  & \bfseries MEL
  & \bfseries BCC
  & \bfseries BKL
  & \bfseries AK
  & \bfseries SCC
  & \bfseries VASC
  & \bfseries DF \\
  \it Class Frequency
  & \it 50.85\,\%
  & \it 17.85\,\%
  & \it 13.12\,\%
  & \it 10.36\,\%
  & \it 3.42\,\%
  & \it 2.48\,\%
  & \it 1.00\,\%
  & \it 0.94\,\% \\
  \hline\hline
  Cross-Entropy
  & 0.8176 & 0.6093 & 0.6382 & 0.4227 & 0.2470 & 0.2521 & 0.3442 & 0.3187 \\
  AdaCos \cite{zhang2019adacos}
  & \textbf{0.8785} & 0.5764 & 0.6687 & 0.4010 & 0.1551 & 0.1080 & 0.1830 & 0.0919 \\
  ArcFace \cite{deng2022arcface}
  & 0.8388 & 0.5931 & 0.6790 & 0.4076 & 0.2674 & 0.1758 & 0.2981 & 0.1319 \\
  Focal Loss \cite{lin2017focalloss}
  & 0.8186 & 0.6273 & 0.6597 & 0.4045 & 0.3011 & 0.2400 & 0.3692 & 0.3055 \\
  CCE Loss \cite{kim2021imbalanced}
  & 0.8326 & 0.6197 & 0.6485 & 0.3995 & 0.2834 & 0.2586 & 0.3141 & 0.3040 \\
  LDAM Loss \cite{cao2019learning}
  & 0.7798 & 0.6345 & \textbf{0.7296} & 0.4002 & 0.2787 & 0.3623 & 0.4210 & 0.4457 \\
  \hline
  CB Loss \cite{cui2019balanced}
  & 0.7870 & 0.6401 & 0.6910 & 0.4321 & 0.2786 & 0.2824 & 0.4250 & 0.4395 \\
  PF Loss \cite{du2021parameter}
  & 0.7248 & 0.6441 & 0.6518 & \textbf{0.4945} & 0.3401 & 0.3454 & 0.4711 & 0.5142 \\
  AWB Loss \cite{yue2022toward}
  & 0.7346 & 0.6710 & 0.7147 & 0.4160 & 0.2695 & 0.2715 & 0.4846 & 0.4901 \\
  DCAL \cite{kim2025long}
  & 0.7862 & 0.6235 & 0.6526 & 0.3530 & 0.3164 & 0.3131 & 0.4295 & 0.3956 \\
  DWCE Loss \eqref{eq:DWCE}
  & 0.7384 & \textbf{0.6832} & 0.6831 & 0.4495 & 0.2888 & 0.3293 & 0.4263 & 0.4241 \\
  \hline
  \bf DAL (Ours)
  & 0.7132 & 0.6532 & 0.6232 & 0.4312 & \textbf{0.4417} & \textbf{0.4125} & \textbf{0.5731} & \textbf{0.6154} \\[-2ex]
  \end{tabular}}

\vspace{1.5ex}

\subtable[APTOS\label{tab:perclass-aptos}]{
  \begin{tabular}{l||c|c|c|c|c}
  \bfseries Method
  & \bfseries No DR
  & \bfseries Moderate
  & \bfseries Mild
  & \bfseries Proliferative DR
  & \bfseries Severe \\
  \it Class Frequency
  & \it 49.29\,\%
  & \it 27.28\,\%
  & \it 10.10\,\%
  & \it 8.06\,\%
  & \it 5.27\,\% \\
  \hline\hline
  Cross-Entropy
  & 0.9845 & 0.8320 & 0.6108 & 0.4475 & 0.1487 \\
  AdaCos \cite{zhang2019adacos}
  & 0.9834 & 0.7401 & 0.6081 & 0.2881 & 0.0556 \\
  ArcFace \cite{deng2022arcface}
  & 0.9850 & 0.7625 & 0.5537 & 0.3688 & 0.3978 \\
  Focal Loss \cite{lin2017focalloss}
  & \textbf{0.9967} & \textbf{0.8570} & 0.6189 & 0.4034 & 0.1795 \\
  CCE Loss \cite{kim2021imbalanced}
  & 0.9825 & 0.8050 & 0.6081 & 0.3503 & 0.1282 \\
  LDAM Loss \cite{cao2019learning}
  & 0.9862 & 0.7570 & 0.5518 & 0.4742 & 0.3616 \\
  \hline
  CB Loss \cite{cui2019balanced}
  & 0.9718 & 0.6381 & 0.7297 & 0.4698 & 0.4192 \\
  PF Loss \cite{du2021parameter}
  & 0.9761 & 0.6530 & 0.7216 & 0.4746 & 0.3798 \\
  AWB Loss \cite{yue2022toward}
  & 0.9801 & 0.7040 & 0.6378 & 0.4983 & 0.4205 \\
  DCAL \cite{kim2025long}
  & 0.9751 & 0.7933 & 0.5315 & 0.4294 & 0.3333 \\
  DWCE Loss \eqref{eq:DWCE}
  & 0.9843 & 0.7610 & 0.6396 & 0.4576 & 0.3504 \\
  \hline
  \bf DAL (Ours)
  & 0.9512 & 0.6321 & \textbf{0.7432} & \textbf{0.5932} & \textbf{0.4723} \\[-2ex]
  \end{tabular}}

\vspace{1.5ex}

\subtable[Hyper-Kvasir\label{tab:perclass-hyperkvasir}]{
  \begin{tabular}{l||c|c|c}
  \bfseries Method
  & \bfseries Head
  & \bfseries Medium
  & \bfseries Tail \\
  \hline\hline
  Cross-Entropy
  & 0.9523 & 0.6578 & 0.0000 \\
  AdaCos \cite{zhang2019adacos}
  & 0.9358 & 0.6387 & 0.0000 \\
  ArcFace \cite{deng2022arcface}
  & 0.9497 & 0.6755 & 0.0000 \\
  Focal Loss \cite{lin2017focalloss}
  & \bf 0.9561 & 0.6822 & 0.0000 \\
  CCE Loss \cite{kim2021imbalanced}
  & 0.9477 & 0.6949 & 0.0000 \\
  LDAM Loss \cite{cao2019learning}
  & 0.9322 & 0.6977 & 0.0231 \\
  \hline
  CB Loss \cite{cui2019balanced}
  & 0.9466 & 0.6712 & 0.0474 \\
  PF Loss \cite{du2021parameter}
  & 0.9360 & 0.6784 & 0.0717 \\
  AWB Loss \cite{yue2022toward}
  & 0.9421 & 0.7050 & 0.0640 \\
  DCAL \cite{kim2025long}
  & 0.9536 & 0.6946 & 0.0000 \\
  DWCE Loss \eqref{eq:DWCE}
  & 0.9503 & 0.6759 & 0.0438 \\
  \hline
  \bf DAL (Ours)
  & 0.9231 & \bf 0.7243 & \bf 0.0946 \\[-2ex]
  \end{tabular}}

\end{center}
\Caption[tab:perclass]{Per-Class Recall}{
    This table shows the per-class recall aka.~true positive rate (TPR$\uparrow$) across our evaluated datasets.
    Classes are sorted by decreasing class frequency.
    Bold numbers highlight best values.
}
\end{table}
\subsubsection{Detailed Analysis}
\tabref{tab:perclass} reports per-class recall.\footnote{Confusion matrices for all loss functions are provided in the supplementary material.} A consistent pattern emerges
across all four datasets: the non-weighted losses such as Cross-Entropy and Focal retain higher recall on the majority classes, whereas DAL stays competitive with the other imbalance-aware baselines there, all of which give up some majority-class recall. More importantly and clinically relevant, the most notable gains appear on the minority and severely under-represented classes, where DAL attains the best recall on the rarest category of every dataset, often by a wide margin. On Hyper-Kvasir the absolute tail recall is low because the
tail classes contain very few test samples, but DAL still attains the highest tail recall among all methods. These improvements highlight the effectiveness of
BAAM and DWCE in addressing strong inter-class imbalance, enabling the model to
better recognize rare disease categories, albeit with some reduction in majority-class recall in line with other imbalance-aware losses.

\subsubsection{Combination with Other Losses}
To analyze the role of ABS from \eqref{eq:adacos} in medical image classification, we combine ABS using our derived $s=4$ and our BAAM \eqref{eq:BAAM} with other loss functions and report the results for ISIC 2019 in \figref{fig:combined}. 
When paired with standard losses such as (unweighted) cross-entropy, focal loss, or LDAM, the angular formulation provides no benefit, but rather decreases the classification accuracies of minority classes -- indicated by a red bar on top -- and impair balanced accuracy (BA) and coefficient of variation (CoV).
In contrast, combined with imbalance-aware losses such as AWB, CB, or DWCE,\footnote{When adding ABS and BAAM to DWCE, we arrive at our proposed DAL.} ABS+BAAM yields the substantial improvements in minority class predictions (green bar) -- at the cost of small reductions for majority classes -- leading to the desired increase of BA and decrease of CoV.

\begin{figure}[t]\centering
\includegraphics[width=0.99\linewidth]{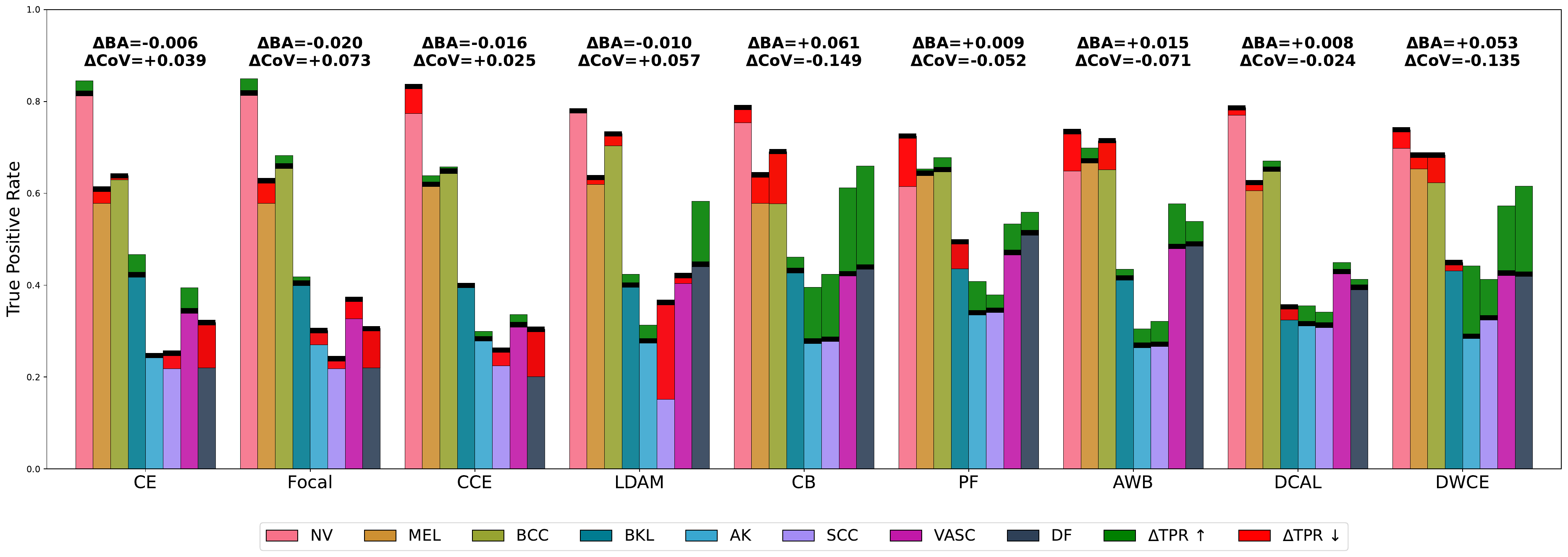}
\Caption[fig:combined]{Combination with Other Losses}{
    For ISIC 2019, we show the per-class recall \raisebox{.5ex}{\rule{.7em}{2pt}} for several state-of-the-art loss functions, and their \textcolor{green}{\bf increases}/\textcolor{red}{\bf reductions} when combined with ABS in \eqref{eq:adacos} using $s = 4$ and BAAM. Changes in balanced accuracy ($\mathrm\Delta$BA$\uparrow$) and coefficient of variation ($\mathrm\Delta$CoV$\downarrow$) are provided numerically.
}
\end{figure}

\begin{figure*}[t]
{
\centering
\subfigure[Cross-Entropy]{\includegraphics[width=.30\textwidth]{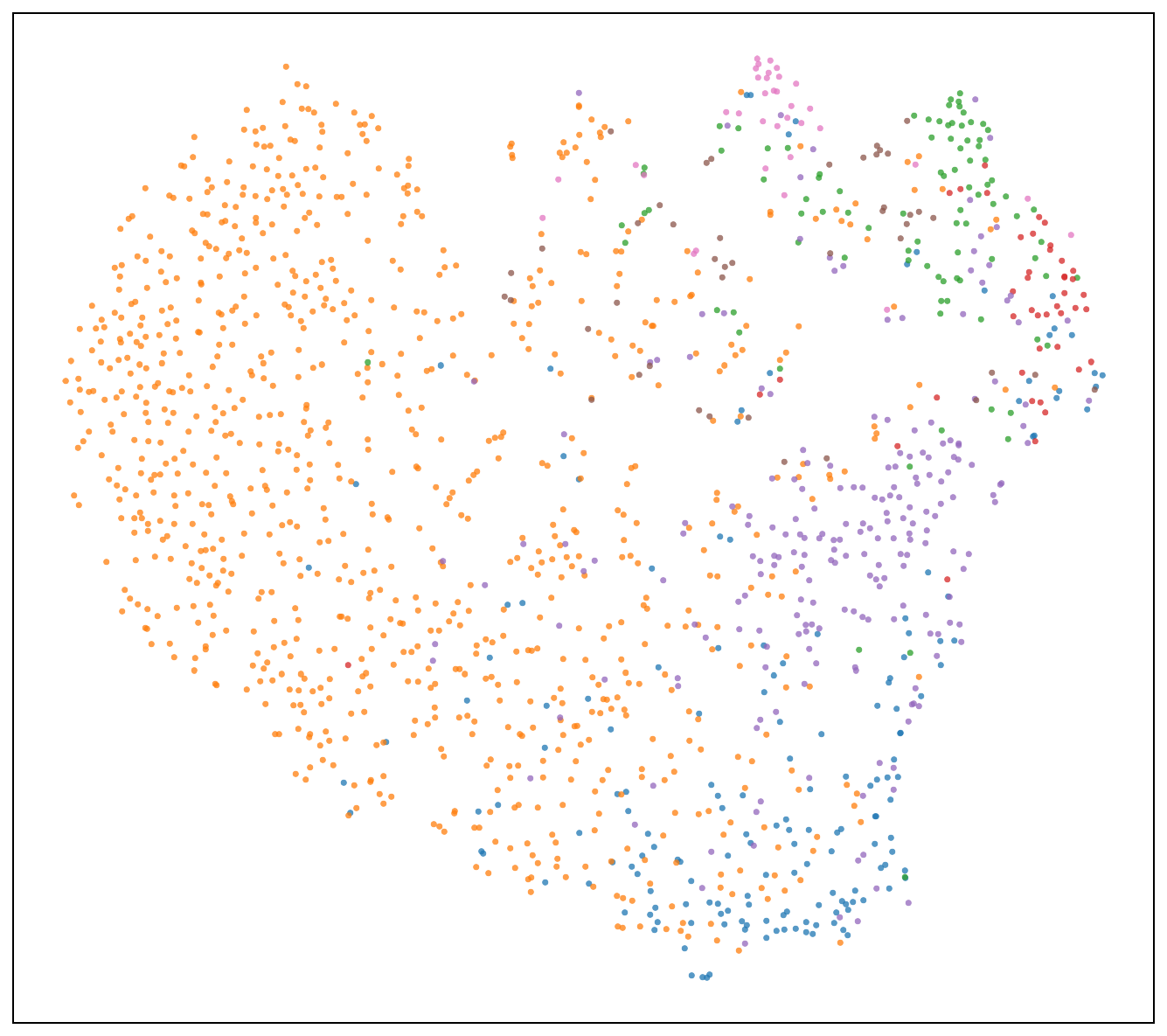}}\quad
\subfigure[Focal]{\includegraphics[width=.30\textwidth]{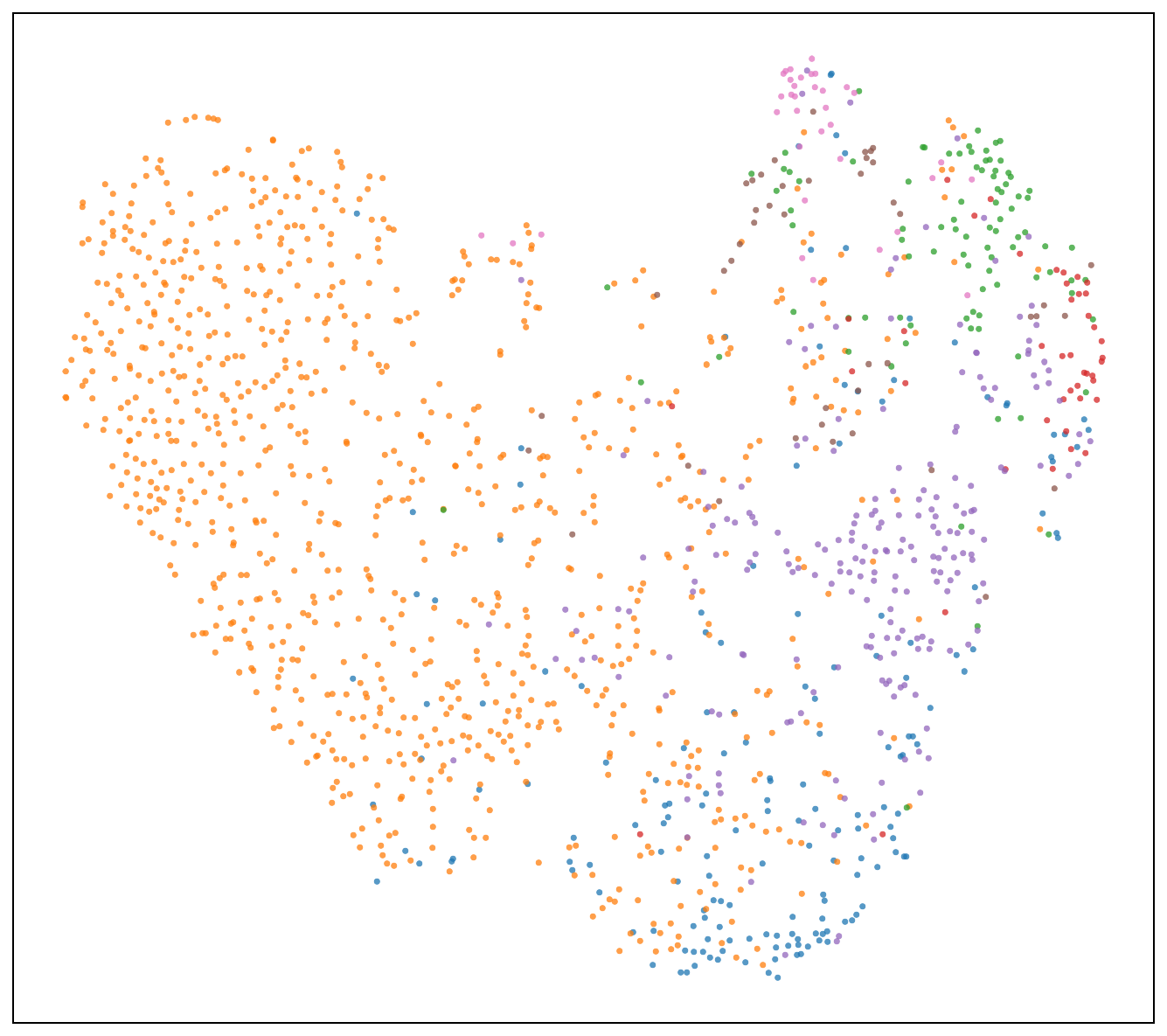}}\quad
\subfigure[LDAM]{\includegraphics[width=.30\textwidth]{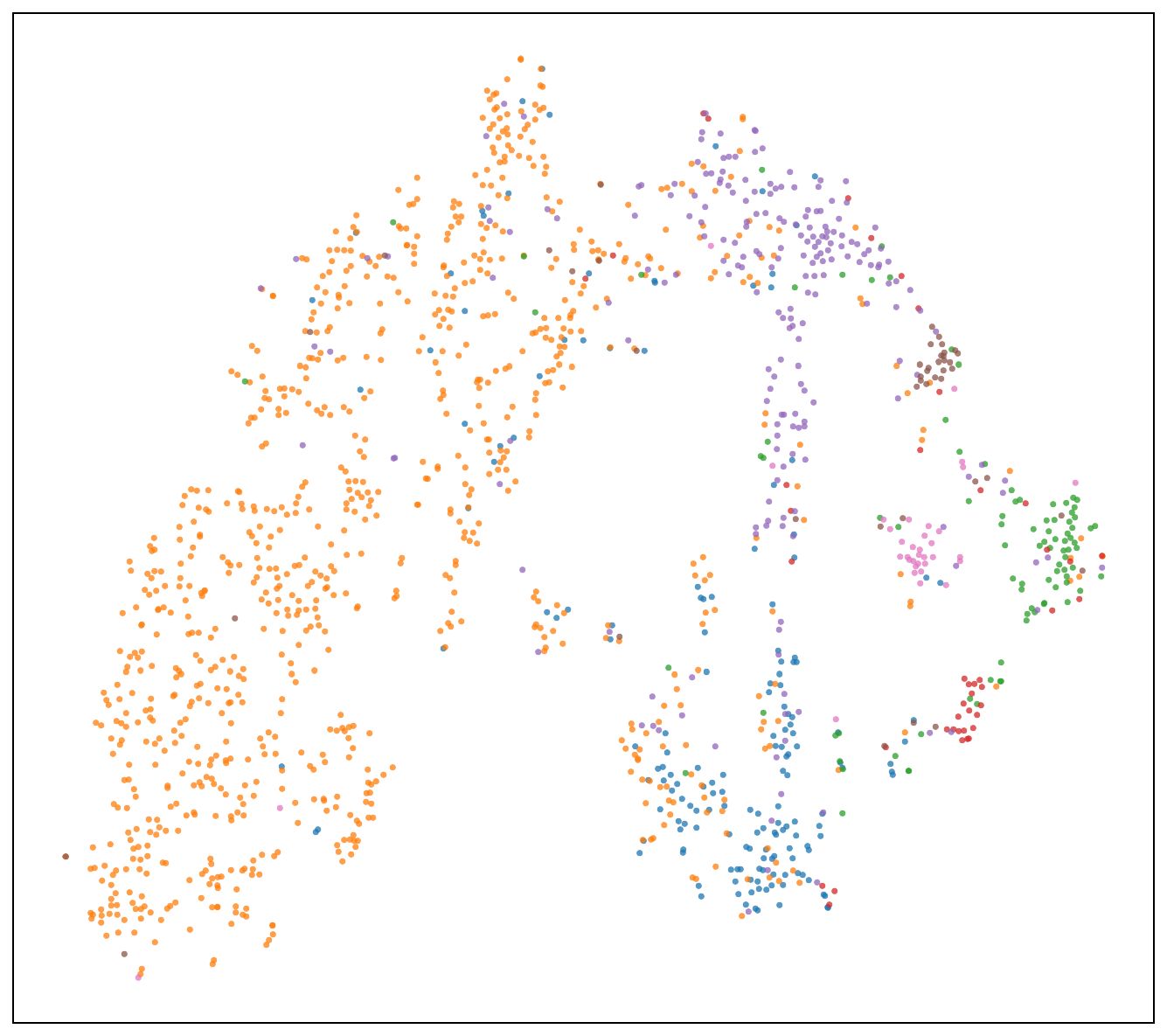}}

\subfigure[CCE]{\includegraphics[width=.30\textwidth]{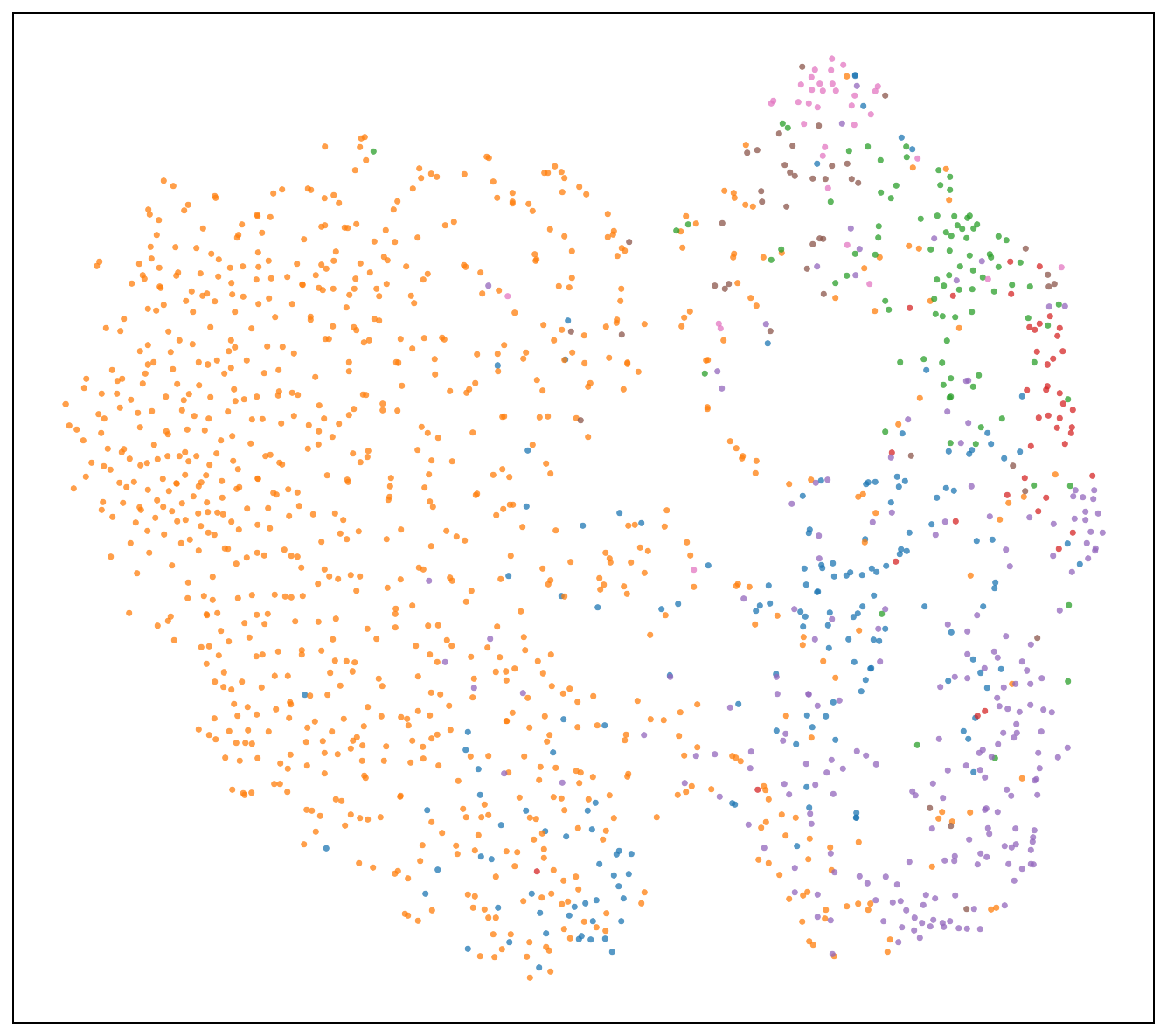}}\quad
\subfigure[CB]{\includegraphics[width=.30\textwidth]{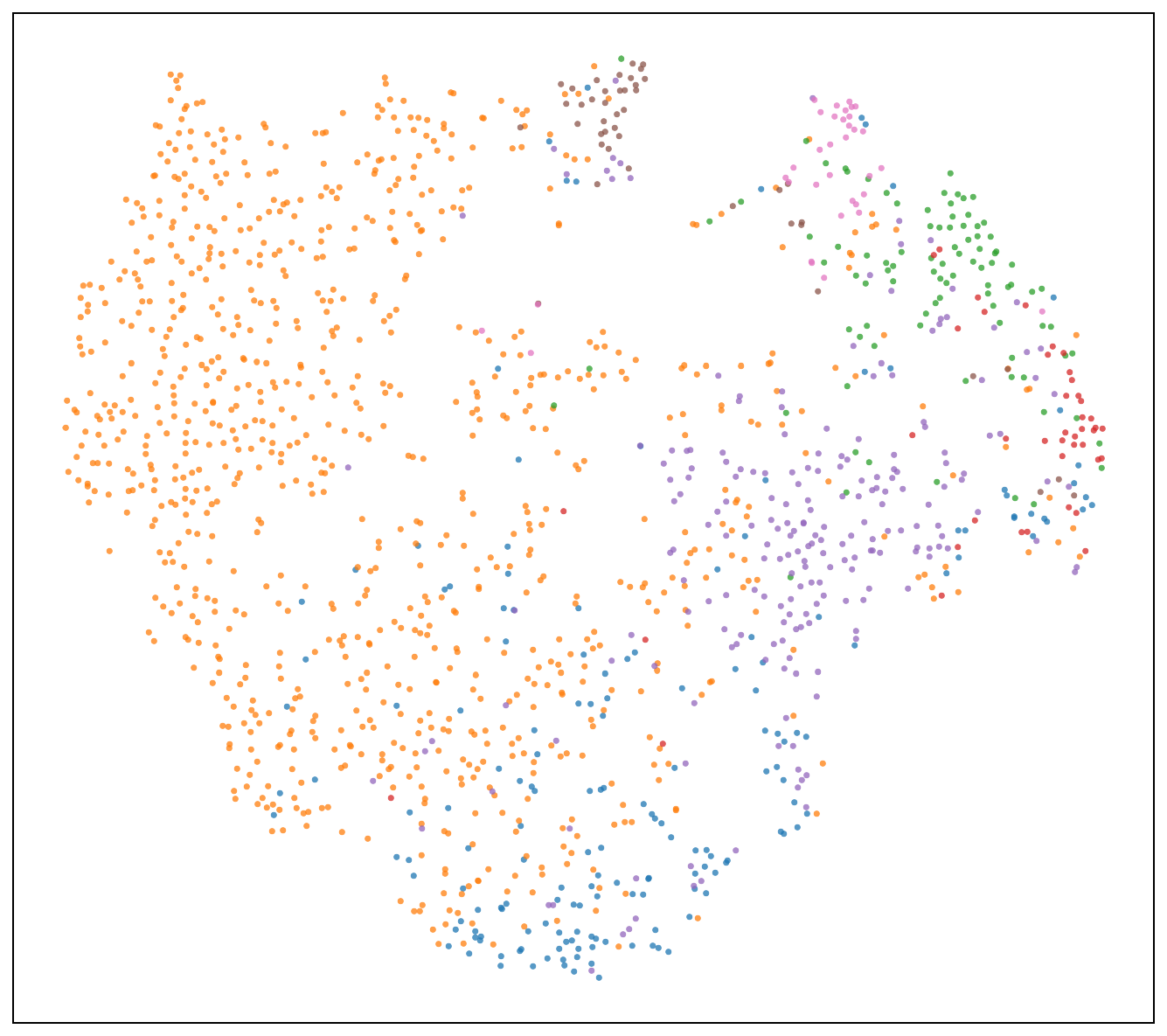}}\quad
\subfigure[DCAL]{\includegraphics[width=.30\textwidth]{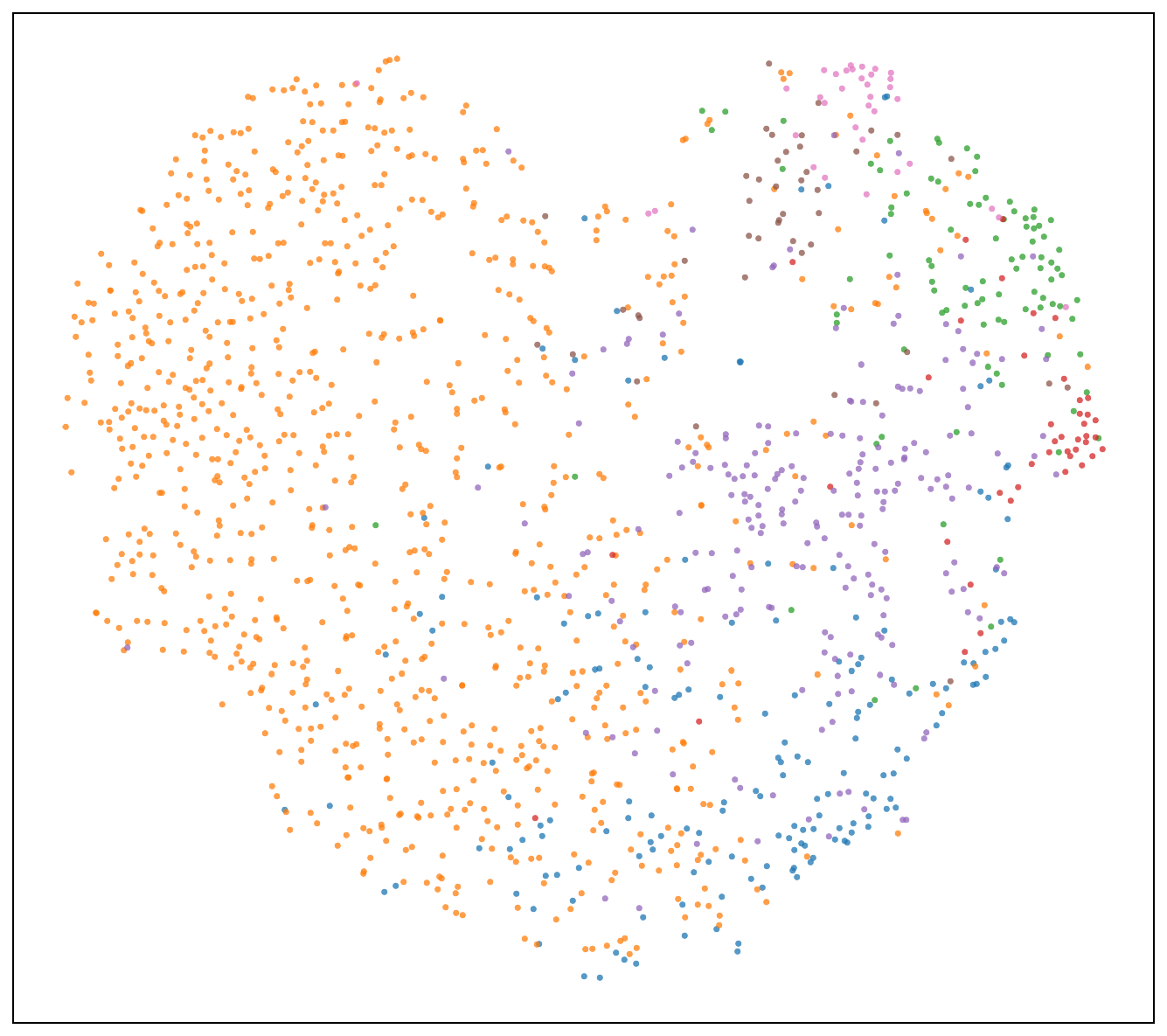}}

\subfigure[AWB]{\includegraphics[width=.30\textwidth]{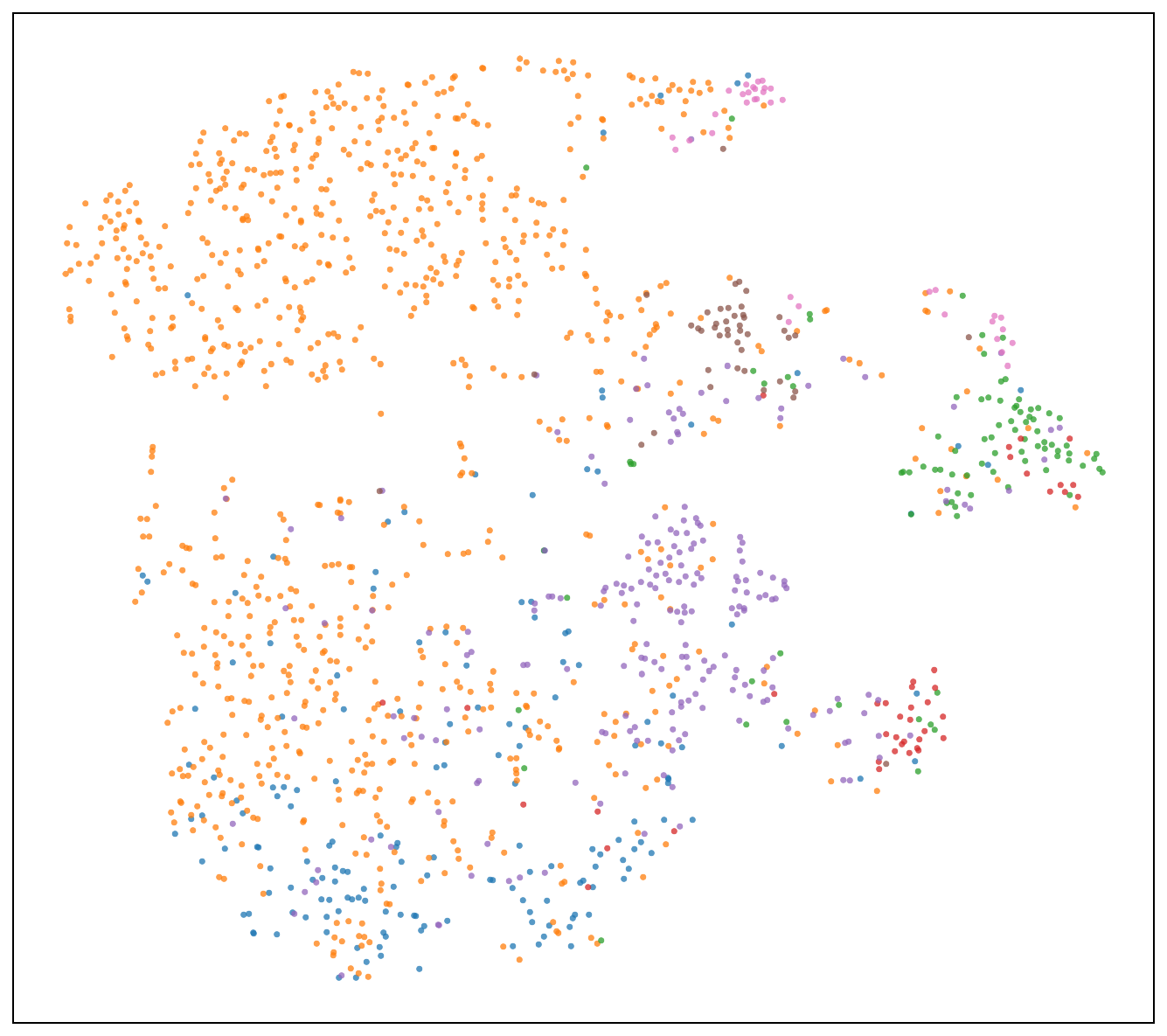}}\quad
\subfigure[PF]{\includegraphics[width=.30\textwidth]{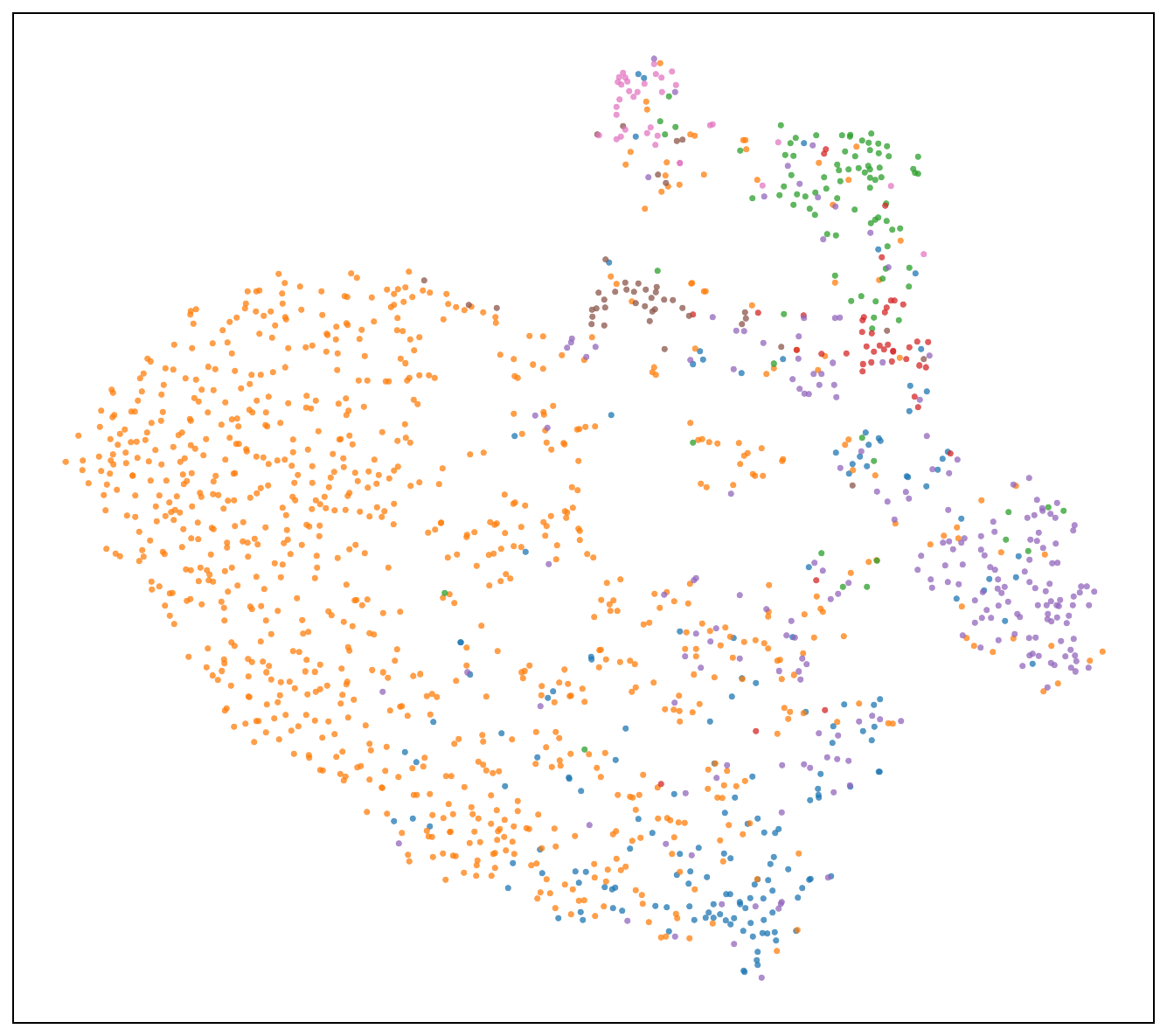}}\quad
\subfigure[DAL (Ours)]{\includegraphics[width=.30\textwidth]{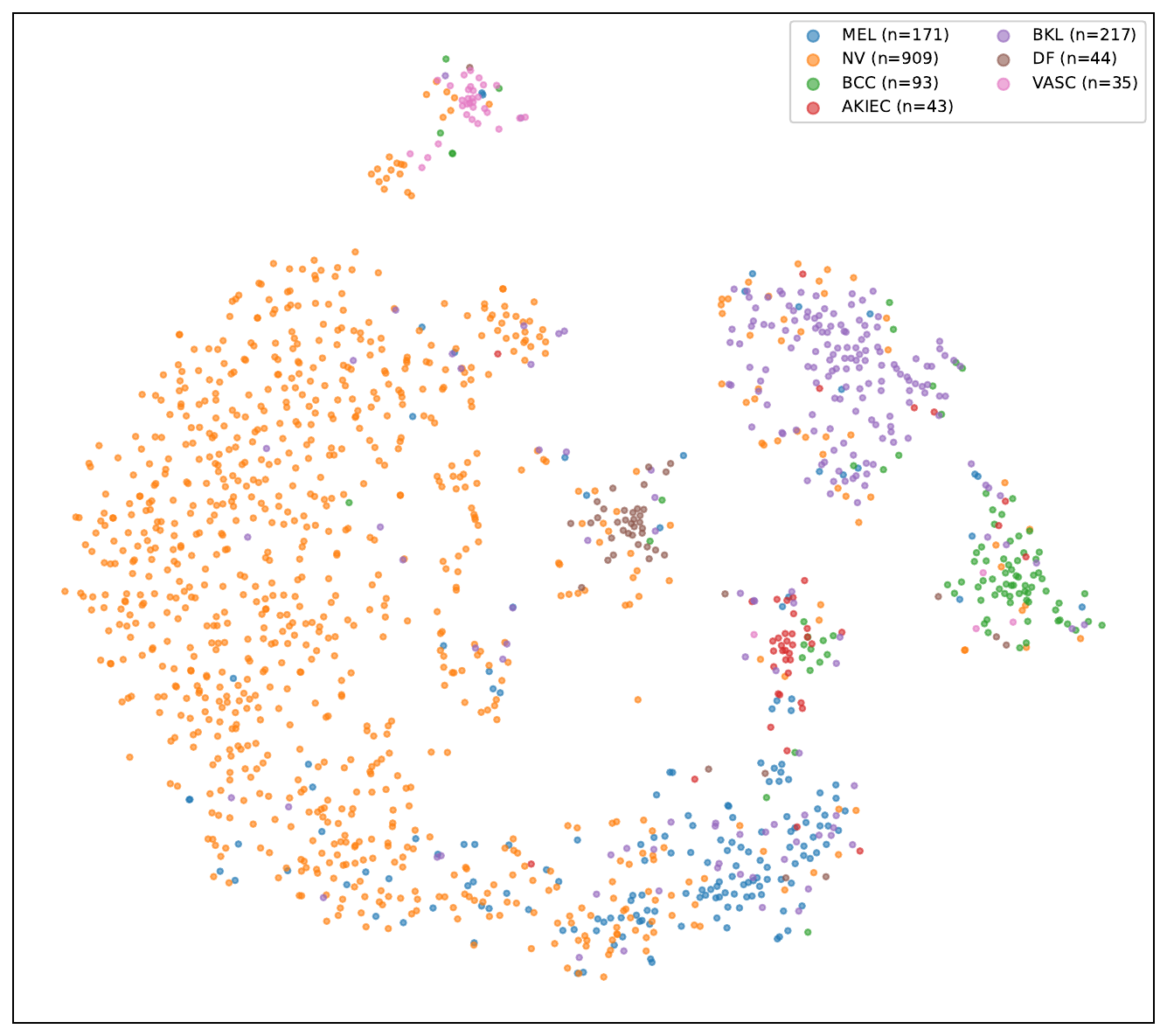}}

}
{\Caption[fig:tsne]{t-SNE Visualization}{We display visualizations of the learned feature space on the ISIC 2018 test set for different loss functions.
}}
\end{figure*}

\subsubsection{Visualization Analysis}
This section presents a comparative analysis of the feature embeddings learned by the baseline losses and the proposed DAL, using t-SNE on the ISIC 2018 test set. For intuitive analysis, we project the feature embeddings into a two-dimensional space, where the embeddings are extracted from the penultimate layer of the backbone. As shown in \figref{fig:tsne}, the
existing methods exhibit indistinct class boundaries and larger within-class dispersion. In contrast, the proposed DAL achieves enhanced inter-class separability and improved intra-class compactness. These visual results demonstrate that DAL enables the model to learn more discriminative feature representations.

\subsubsection{Generalization Across Architectures}
\label{sec:architecture-generalization}

\begin{table}[t]
\begin{center}
\scriptsize
\begin{tabular}{l||c|c|c||c|c|c}
\multirow{2}{*}{\bfseries Method}
& \multicolumn{3}{c||}{\bfseries ISIC 2018} & \multicolumn{3}{c}{\bfseries APTOS 2019} \\
\cline{2-4}\cline{5-7}
& \bfseries Acc$\uparrow$ & \bfseries BAcc$\uparrow$ & \bfseries F1$\uparrow$ 
& \bfseries Acc$\uparrow$ & \bfseries BAcc$\uparrow$ & \bfseries F1$\uparrow$ \\
\hline
\multicolumn{7}{l}{\textit{DenseNet121}} \\
LDAM-DRW \cite{cao2019learning}
& --- & 0.6876 & 0.6820 
& --- & 0.6217 & 0.6225 \\
DWB \cite{fernando2021dynamically}
& --- & 0.6204 & 0.6826 
& --- & 0.6555 & 0.6417 \\
MAAW \cite{roy2023margin}
& --- & 0.7060 & 0.7019 
& --- & 0.6733 & 0.6701 \\
\bf DAL (Ours) 
& 0.8448 & \textbf{0.7685} & \textbf{0.7480} 
& 0.8172 & \textbf{0.6934} & \textbf{0.6887} \\
\hline
\multicolumn{7}{l}{\textit{ResNet50}} \\
BPaCo \cite{cai2024bpaco}
& 0.8650 & 0.7580 & 0.7600 
& \textbf{0.8340} & 0.6650 & 0.6340 \\
CBTS \cite{zhao2025constructing}
& 0.8490 & \textbf{0.8010} & 0.7340 
& 0.8000 & 0.6440 & 0.6360 \\
CACL \cite{deng2026class}
& \textbf{0.8840} & 0.7860 & 0.7980 
& \textbf{0.8340} & 0.6650 & 0.6770 \\
\bf DAL (Ours) 
& 0.8577 & 0.7990 & \textbf{0.7990} 
& 0.8217 & \textbf{0.6801} & \textbf{0.6853} \\
\end{tabular}
\end{center}

\Caption[tab:architecture-comparison]{Performance Across Backbone Architectures}{
    We report 5-fold cross-validation performance of our method (DAL) with alternative backbone
    architectures, compared against state-of-the-art imbalanced-classification
    methods reported with the same backbones on ISIC 2018 and APTOS 2019.
    Best results within each architecture group are highlighted in bold;
    entries marked ``---'' were not reported by the corresponding works.
}
\end{table}

Our formulation is independent of the backbone network: the scale and margin
bounds depend only on the number of classes and the negative-angle
statistics observed during training, both of which are available for any
embedding architecture. To demonstrate this, we apply DAL with two widely
used backbones, DenseNet121 and ResNet50, and compare against
state-of-the-art imbalanced-classification methods that report results with
the same backbones (\tabref{tab:architecture-comparison}). For the competing
methods, we report the numbers stated in \cite{roy2023margin} and \cite{deng2026class} respectively. Across both architectures, DAL achieves the best balanced accuracy and F1 on APTOS and the best F1 on ISIC 2018, while remaining competitive on overall accuracy. Where competing methods attain higher raw accuracy (e.g., CACL on ISIC 2018 with ResNet50), this reflects stronger majority-class performance; DAL's consistent advantage on the imbalance-sensitive metrics indicates more equitable per-class behavior, in line with our objective of a reliable classifier under severe imbalance. These results confirm that the method transfers across backbone architectures without modification.

\paragraph{An Important Note on ISIC Evaluation Protocols.}
The ISIC 2018 accuracy and balanced-accuracy values in
\tabref{tab:architecture-comparison} are notably higher than those reported
in our main experiments (\tabref{tab:test-results}). This discrepancy is not
due to the method but to the evaluation protocol: the competing methods, from
whose papers we take the reported numbers, do not evaluate on the official
ISIC 2018 test set, which uses a subject-independent split in which no
patient appears in both training and test partitions. Evaluating on
random splits, where images from the same patient may appear in both
partitions, inflates performance because the model can exploit
patient-specific features rather than learning generalizable
class-discriminative representations. To enable a fair comparison with the
numbers reported by competing works, the results in
\tabref{tab:architecture-comparison} follow the same (non-official) splitting
protocol used in those works. We emphasize that our main results
(\tabref{tab:test-results}) use the official subject-independent test set,
which is the more rigorous and clinically meaningful evaluation, and we
encourage future work on ISIC datasets to report on this split to avoid
optimistically biased estimates.
\subsection{Ablation Study}
\label{sec:ablation}
\begin{table}[t]

\begin{center}
\scriptsize
\begin{tabular}{l||c|c||c|c|c|c|c|c}
\bfseries Method & \bfseries $s$ & \bfseries $m$ 
& \bfseries Acc & \bfseries F1 & \bfseries BAcc
& \bfseries IBA$_1$
& \bfseries CoV & \bfseries p-value \\
\hline
ArcFace & 4  & 0.0 & 0.7553 & 0.6458 & 0.7238 & 0.5352 & 0.1068 & $\mathbf{4.214\mathrm{e}{-3}}$ \\
ArcFace & 4 & 0.1 & 0.7665 & 0.6560 & 0.7103 & 0.5101 & 0.1138 & \( \mathbf{1.615 \mathrm{e}{-3}} \) \\
ArcFace & 4  & 0.2 & 0.7595 & 0.6601 & 0.7060 & 0.5032 & 0.1223 & $\mathbf{1.211\mathrm{e}{-3}}$ \\
ArcFace & 4  & 0.3 & 0.7684 & 0.6618 & 0.6919 & 0.4992 & 0.1049 & $\mathbf{5.213\mathrm{e}{-4}}$ \\
\hline
ArcFace & 8  & 0.1 & 0.7559 & 0.6604 & 0.7113 & 0.5271 & \textbf{0.0957} & $\mathbf{3.112\mathrm{e}{-3}}$ \\
ArcFace & 16 & 0.3 & \bf 0.7698 & \textbf{0.6651} & 0.7035 & 0.5090 & 0.1038 & $\mathbf{1.442\mathrm{e}{-3}}$ \\
\hline
AdaCos (fixed) & 2.56 & 0.0 & 0.7251 & 0.6162 & 0.7204 & 0.5316 & 0.1050 & $\mathbf{3.418\mathrm{e}{-3}}$ \\
AdaCos & dyn & 0.0 & 0.7163 & 0.6044 & 0.7018 & 0.5034   & 0.1348 & $\mathbf{1.712\mathrm{e}{-3}}$ \\
\hline
\textbf{DAL (Ours)} & 4 & BAAM & 0.7443 & 0.6490 & \textbf{0.7348} & \textbf{0.5470} & 0.0972 & ---\\
\end{tabular}
\end{center}

\Caption[tab:isic2018hyperparameters]{Ablation Study}{
    We report the test-set performance for ArcFace models under different combinations of
    scale $s$ and margin $m$. All experiments are run including dynamic weighted cross-entropy (DWCE).
    Metrics include overall accuracy (Acc), macro F1-score (F1), balanced
    accuracy (BAcc), the imbalance-aware index (IBA) at $\alpha=1$, and the coefficient of variation (CoV), as well as p-values for a paired t-test against our method.
}

\end{table}
Here, we ablate the angular-softmax scale $s$ and margin $m$, alongside two
AdaCos variants and our method (DAL).\footnote{The supplementary material provides further ablations: a batch-size sweep on ISIC~2018 and a scale sensitivity sweep on APTOS.}
All configurations in \tabref{tab:isic2018hyperparameters} are trained with DWCE and evaluated on the official ISIC2018 test set.

\subsubsection{Scale--Margin Interaction in ArcFace}
The optimal margin depends on the scale, in line with the admissible range we derive in \secref{sec:scale}. At our derived scale $s = 4$, balanced accuracy is highest with no margin: the admissible window is narrow here, so a fixed positive margin has little room to help. As the scale grows, the window widens and the best-performing margin increases with it, matching our bound $m_{\max}$ growing with $s$ (\secref{sec:Margin}). Despite needing larger margins, the higher scales never overtake $s = 4$ on balanced accuracy, although F1 score does improve a bit. Overall, DAL achieves the best results on the imbalance-aware metrics central to our setting while staying competitive on F1 and overall accuracy.

\subsubsection{Comparison with AdaCos and DAL}
Fixed-scale AdaCos with $s = 2.56$ according to \eqref{eq:adacos-s} attains a balanced accuracy comparable to ArcFace with $s = 4$, while the dynamic-scale variant is lower. This apparent parity is misleading, however: AdaCos reaches a
comparable balanced accuracy only at a clear cost in overall accuracy and F1, both of which fall below ArcFace at $s = 4$ and below DAL. DAL attains the highest balanced accuracy in \tabref{tab:isic2018hyperparameters}, exceeding
every fixed $(s, m)$ ArcFace configuration and both AdaCos variants, while keeping reasonable accuracy and low CoV. All variants differ significantly from DAL under a paired $t$-test ($p < 0.01$ in every case).
\section{Conclusion}
In this paper, we investigated the usability of \emph{dynamic angular-margin losses} (DAL) for improving predictions of minority classes in imbalanced medical image classification tasks.
First, we analyzed the scale parameter $s$ of \emph{angle-based softmax} (ABS) and showed that the basic assumption from \citet{zhang2019adacos} does not translate to medical datasets with small numbers of classes.
We derived theoretical upper and lower bounds for $s$ that work much better with fewer classes, as well as an upper bound for the margin $m$.
DAL couples representation-level adaptation through \emph{batch-adaptive angular margins} (BAAM) with loss-level balancing via \emph{dynamically-weighted cross-entropy} (DWCE), which  encourages minority classes to produce more compact clusters and receive stronger optimization signals, and prevents majority classes  from dominating the shared feature space or gradient updates. 
This joint mechanism provides a robust and reliable approach to mitigate severe class imbalance.

We ran experiments on four imbalanced medical image classification tasks and showed that DAL improves classification accuracy for underrepresented classes, and provides a more fair model as measured by the coefficient of variation. In general, our method maintains stable recall across the full class spectrum, indicating that the proposed framework yields robust class-wise generalization and effectively handles long-tailed medical image distributions.
We also highlighted the importance of ABS by adding it to existing loss functions, and observed performance improvements when loss functions explicitly balance class distributions. Notably, focal loss, which is well-adopted for imbalanced classification tasks, rather focused on improving majority classes, and also did not benefit from ABS. Finally, simply relying on aggregate metrics, \eg, accuracy or F1 can lead to misleading conclusions when the classification of minority classes is of importance.

\bibliography{sections/References,sections/Publications,sections/bibliography}
\end{document}


\maketitle

\section{Effect of Batch Composition on BAAM}
\label{sec:ablation-batch}

\begin{figure}[b]\centering
\includegraphics[width=0.75\linewidth]{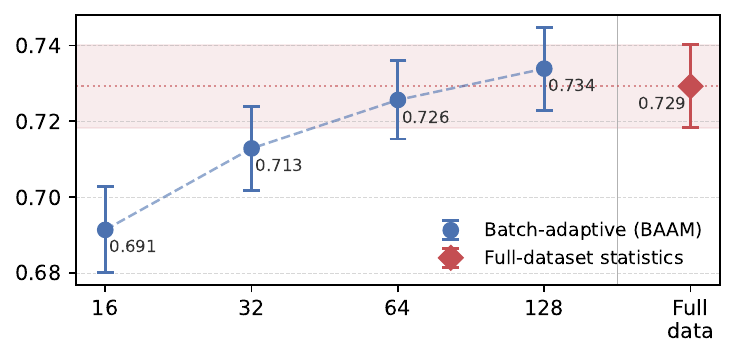}
\Caption[fig:batch]{Batch-size Sensitivity on ISIC 2018}{We show BAcc against batch size $N$, computed as the mean over five folds with the 95\% Confidence Intervals.}
\end{figure}

BAAM sets its margins from the class counts in each mini-batch, and those counts
can be noisy. In \figref{fig:batch}, we test the stability of the method with respect to batch size on
ISIC~2018, and compare against a variant whose margins come from the class
frequencies of the whole training set. BAcc improves as the batch grows, since
larger batches estimate the dataset class counts more closely and give more
stable margins. From a batch size of 64 onwards the batch-based margins are
on par with the full-dataset margins, and the confidence intervals overlap. At
our operating point, using per-batch counts costs no performance against dataset
counts, while it keeps the margins tied to the batch the update is computed on.

\section{Scale Sensitivity}

\begin{figure}[t]\centering
\includegraphics[width=0.65\linewidth]{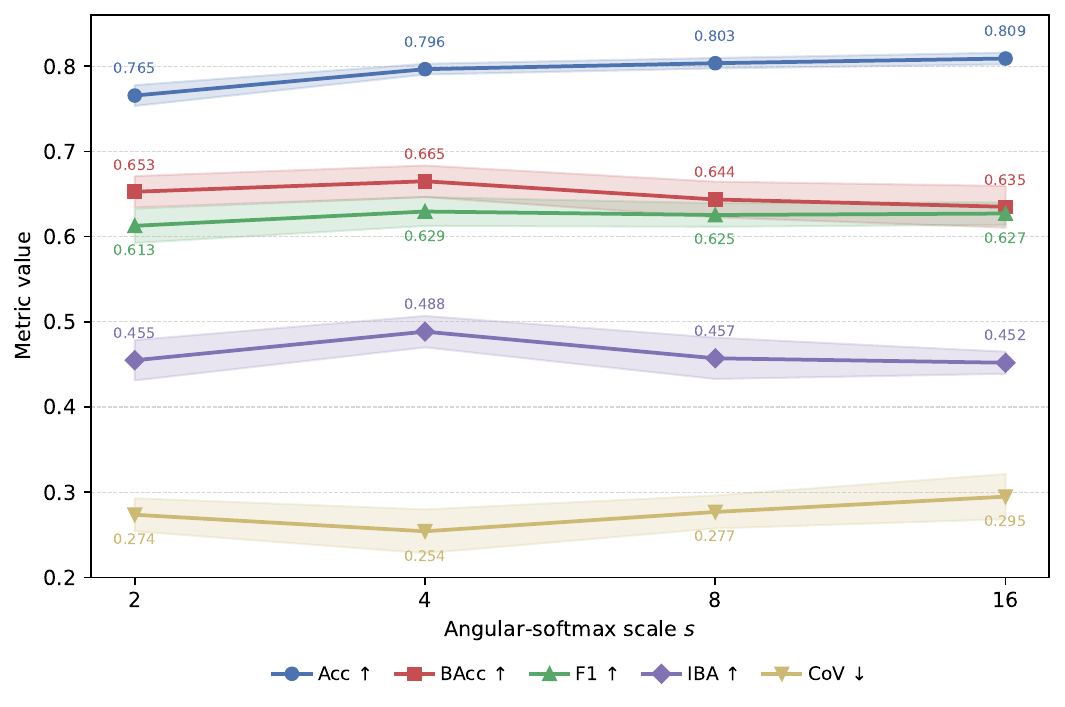}
\Caption[fig:scale-aptos]{Scale Sensitivity on APTOS}{We plot metric values against the angular-softmax scale $s$, and report means over five runs, the shaded regions indicate 95\% Confidence Intervals.}
\end{figure}

\figref{fig:scale-aptos} reports a sensitivity sweep over the angular-softmax
scale on APTOS, with $s \in \{2, 4, 8, 16\}$ and five runs per setting. Every
imbalance-sensitive metric peaks at or near $s{=}4$: BAcc, F1 and IBA are
highest there, and CoV is lowest. Accuracy is the only metric that rises
monotonically with $s$, which is consistent with a larger scale sharpening the
softmax in favor of the majority classes. The same behavior appears on
ISIC~2018, where BAcc decreases once $s$ grows past 4. The confidence bands of
neighboring settings overlap, so we read this as an ordering of the point
estimates rather than a significant difference between them. We report the sweep
as a post-hoc check on the behavior of $s$, not as the procedure we used to
select it.

\section{Angle Distributions}

\figref{fig:angles-supp} shows the distribution of the negative-class ArcFace
angles $\theta_{i,j}$ over the course of training, on the training sets of
APTOS, Hyper-Kvasir and ISIC~2019. These plots complement the corresponding
figure for ISIC~2018 in the main paper. While not identical to the distributions of ISIC~2018 as reported in the main paper, ranges are similar, and the bounds estimated on ISIC~2018 can be roughly translated to other medical datasets.

\begin{figure*}[p]
{
\centering
\subfigure[APTOS]{\includegraphics[width=.48\textwidth]{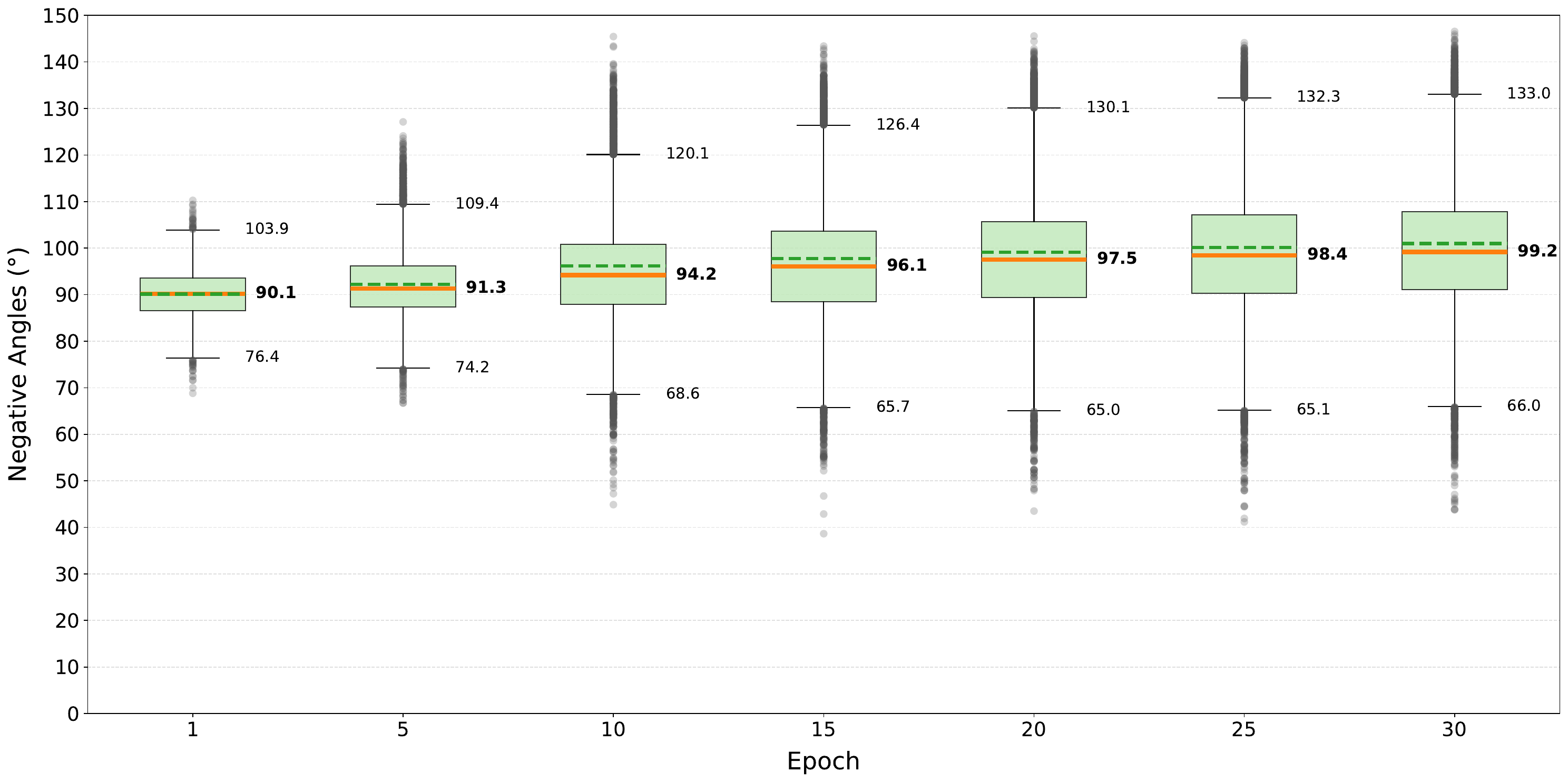}}\hfill
\subfigure[Hyper-Kvasir]{\includegraphics[width=.48\textwidth]{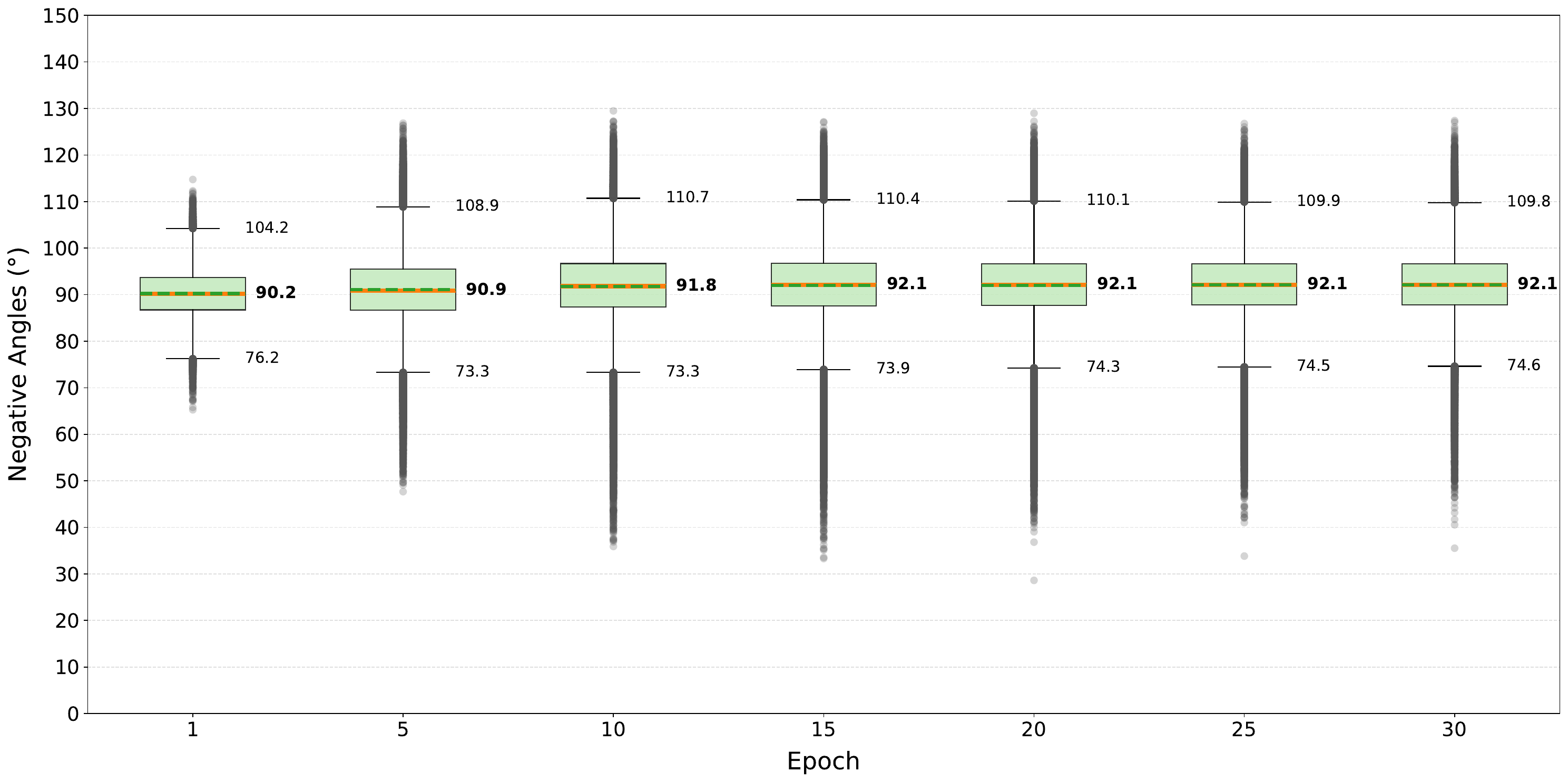}}\\[1ex]
\subfigure[ISIC 2019]{\includegraphics[width=.48\textwidth]{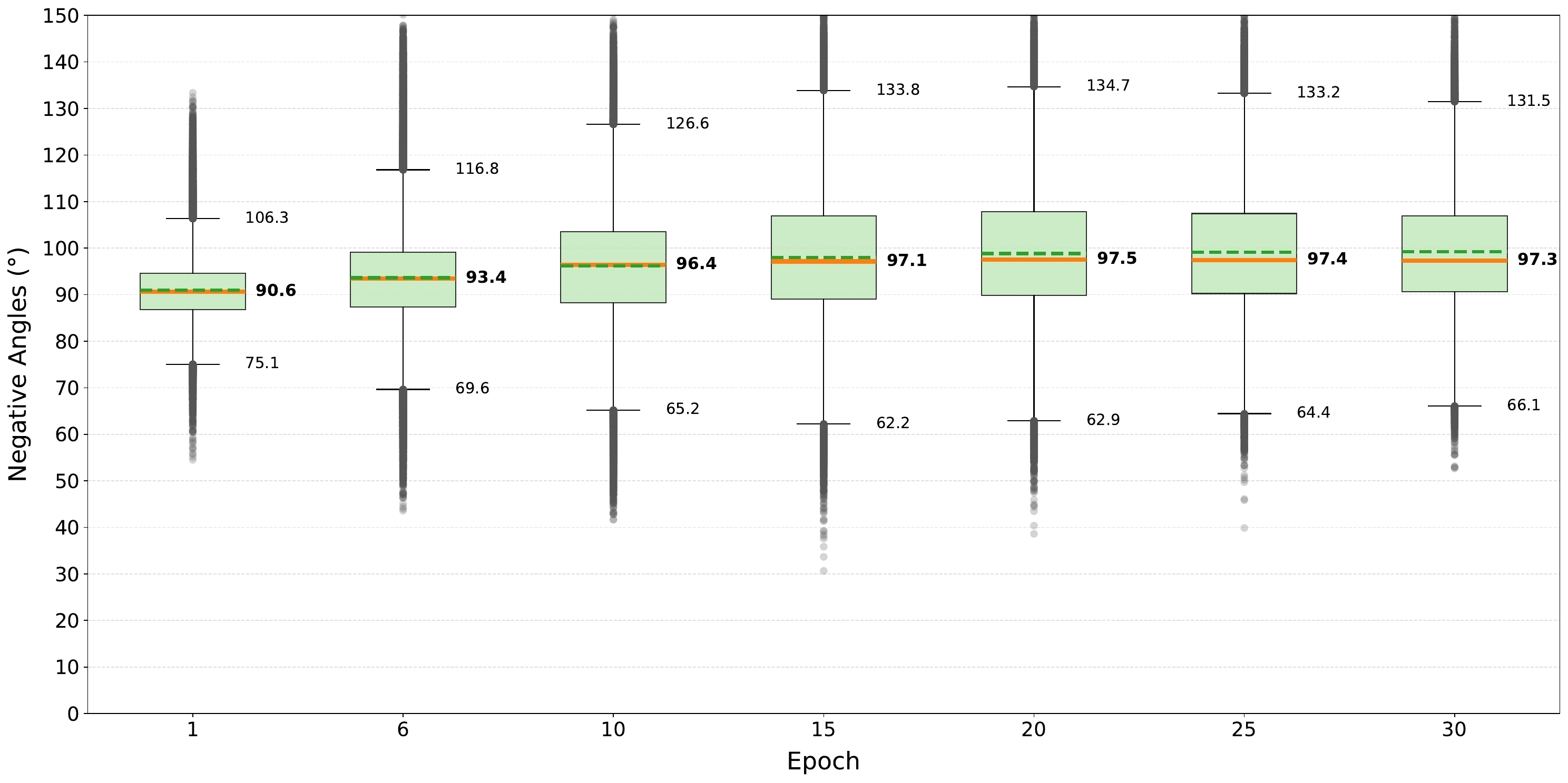}}
\par}
{\Caption[fig:angles-supp]{Negative Angles per Epoch}{We show box plots of the distribution of ArcFace angles $\theta_{i,j}$ for negative classes during training, on the APTOS, Hyper-Kvasir and ISIC 2019 training sets.}}
\end{figure*}

\section{Uncertainty Estimates}

\tabref{tab:uncertainty} repeats the comparison of Table 1 of the main paper with
uncertainty estimates attached to every entry. We report the mean over five
folds together with the corresponding 95\% confidence interval, for all methods,
datasets and metrics.
In general, confidence bounds do not overlap, indicating statistically significant improvements of DAL over competitors for bias-aware BA and IBA$_1$ metrics, and often also for CoV.
\begin{table}[p]
\setlength{\tabcolsep}{2pt}
\begin{center}
\scriptsize

\begin{tabular}{@{}l||c|c|c|c|c||c|c|c|c|c@{}}
\multirow{2}{*}{\bfseries Method}
 & \multicolumn{5}{c||}{ISIC 2018} & \multicolumn{5}{c}{ISIC 2019} \\
\cline{2-11}
 & \bfseries Acc$\uparrow$
 & \bfseries F1$\uparrow$
 & \bfseries BA$\uparrow$
 & \bfseries IBA$_1$$\uparrow$
 & \bfseries CoV$\downarrow$
 & \bfseries Acc$\uparrow$
 & \bfseries F1$\uparrow$
 & \bfseries BA$\uparrow$
 & \bfseries IBA$_1$$\uparrow$
 & \bfseries CoV$\downarrow$ \\
\hline
Cross-Entropy
 & \ms{0.778}{0.003} & \ms{0.618}{0.008} & \ms{0.602}{0.009}
 & \ms{0.389}{0.010} & \ms{0.261}{0.029}
 & \ms{0.638}{0.008} & \ms{0.463}{0.014} & \ms{0.456}{0.013}
 & \ms{0.255}{0.010} & \ms{0.428}{0.031} \\
AdaCos \cite{zhang2019adacos}
 & \ms{0.774}{0.011} & \ms{0.590}{0.013} & \ms{0.590}{0.014}
 & \ms{0.410}{0.014} & \ms{0.291}{0.038}
 & \ms{0.639}{0.011} & \ms{0.374}{0.014} & \ms{0.383}{0.018}
 & \ms{0.224}{0.019} & \ms{0.725}{0.039} \\
ArcFace \cite{deng2022arcface}
 & \ms{0.786}{0.008} & \ms{0.638}{0.009} & \ms{0.641}{0.010}
 & \ms{0.419}{0.010} & \ms{0.210}{0.025}
 & \ms{0.643}{0.011} & \ms{0.452}{0.010} & \ms{0.424}{0.011}
 & \ms{0.242}{0.013} & \ms{0.561}{0.033} \\
Focal Loss \cite{lin2017focalloss}
 & \ms{0.772}{0.006} & \ms{0.620}{0.008} & \ms{0.610}{0.012}
 & \ms{0.397}{0.015} & \ms{0.247}{0.026}
 & \ms{0.647}{0.006} & \ms{0.479}{0.008} & \ms{0.469}{0.010}
 & \ms{0.263}{0.007} & \ms{0.419}{0.023} \\
CCE Loss \cite{kim2021imbalanced}
 & \ms{0.774}{0.013} & \ms{0.616}{0.019} & \ms{0.608}{0.018}
 & \ms{0.399}{0.018} & \ms{0.234}{0.024}
 & \ms{0.647}{0.006} & \ms{0.475}{0.003} & \ms{0.458}{0.004}
 & \ms{0.257}{0.003} & \ms{0.438}{0.017} \\
LDAM Loss \cite{cao2019learning}
 & \ms{\textbf{0.786}}{0.013} & \ms{0.660}{0.017} & \ms{0.668}{0.035}
 & \ms{0.455}{0.039} & \ms{0.204}{0.081}
 & \ms{0.632}{0.007} & \ms{0.476}{0.009} & \ms{0.480}{0.007}
 & \ms{0.287}{0.010} & \ms{0.339}{0.034} \\
\hline
CB Loss \cite{cui2019balanced}
 & \ms{0.776}{0.020} & \ms{\textbf{0.668}}{0.019} & \ms{0.673}{0.015}
 & \ms{0.466}{0.018} & \ms{0.130}{0.039}
 & \ms{\textbf{0.647}}{0.005} & \ms{0.511}{0.012} & \ms{0.497}{0.019}
 & \ms{0.288}{0.013} & \ms{0.359}{0.054} \\
PF Loss \cite{du2021parameter}
 & \ms{0.735}{0.014} & \ms{0.647}{0.009} & \ms{0.694}{0.011}
 & \ms{0.494}{0.016} & \ms{\textbf{0.093}}{0.007}
 & \ms{0.631}{0.009} & \ms{0.503}{0.012} & \ms{0.523}{0.015}
 & \ms{0.304}{0.012} & \ms{0.253}{0.040} \\
AWB Loss \cite{yue2022toward}
 & \ms{0.731}{0.045} & \ms{0.637}{0.024} & \ms{0.650}{0.011}
 & \ms{0.438}{0.013} & \ms{0.137}{0.026}
 & \ms{0.629}{0.018} & \ms{\textbf{0.513}}{0.015} & \ms{0.505}{0.012}
 & \ms{0.294}{0.011} & \ms{0.344}{0.020} \\
DCAL \cite{kim2025long}
 & \ms{0.769}{0.009} & \ms{0.631}{0.021} & \ms{0.636}{0.032}
 & \ms{0.427}{0.036} & \ms{0.205}{0.046}
 & \ms{0.633}{0.008} & \ms{0.489}{0.004} & \ms{0.481}{0.007}
 & \ms{0.271}{0.006} & \ms{0.346}{0.038} \\
DWCE Loss 
 & \ms{0.763}{0.023} & \ms{0.661}{0.022} & \ms{0.679}{0.025}
 & \ms{0.472}{0.028} & \ms{0.134}{0.039}
 & \ms{0.639}{0.016} & \ms{0.509}{0.014} & \ms{0.503}{0.017}
 & \ms{0.289}{0.020} & \ms{0.323}{0.051} \\
\hline
DAL (Ours)
 & \ms{0.744}{0.024} & \ms{0.649}{0.012} & \ms{\textbf{0.735}}{0.011}
 & \ms{\textbf{0.547}}{0.016} & \ms{0.097}{0.016}
 & \ms{0.621}{0.015} & \ms{0.506}{0.009} & \ms{\textbf{0.552}}{0.011}
 & \ms{\textbf{0.334}}{0.009} & \ms{\textbf{0.188}}{0.040} \\

\end{tabular}

\vspace{1.5ex}

\begin{tabular}{@{}l||c|c|c|c|c||c|c|c|c|c@{}}
\multirow{2}{*}{\bfseries Method}
 & \multicolumn{5}{c||}{APTOS} & \multicolumn{5}{c}{Hyper-Kvasir} \\
\cline{2-11}
 & \bfseries Acc$\uparrow$
 & \bfseries F1$\uparrow$
 & \bfseries BA$\uparrow$
 & \bfseries IBA$_1$$\uparrow$
 & \bfseries CoV$\downarrow$
 & \bfseries Acc$\uparrow$
 & \bfseries F1$\uparrow$
 & \bfseries BA$\uparrow$
 & \bfseries IBA$_1$$\uparrow$
 & \bfseries CoV$\downarrow$ \\
\hline\hline
Cross-Entropy
 & \ms{0.818}{0.010} & \ms{0.616}{0.017} & \ms{0.605}{0.010}
 & \ms{0.460}{0.012} & \ms{0.484}{0.029}
 & \ms{\textbf{0.893}}{0.006} & \ms{0.575}{0.009} & \ms{0.578}{0.009}
 & \ms{0.511}{0.010} & \ms{0.728}{0.011} \\
AdaCos \cite{zhang2019adacos}
 & \ms{0.772}{0.012} & \ms{0.525}{0.021} & \ms{0.529}{0.016}
 & \ms{0.399}{0.017} & \ms{0.614}{0.037}
 & \ms{0.881}{0.008} & \ms{0.571}{0.008} & \ms{0.569}{0.007}
 & \ms{0.500}{0.007} & \ms{0.721}{0.010} \\
ArcFace \cite{deng2022arcface}
 & \ms{0.800}{0.009} & \ms{0.626}{0.013} & \ms{0.614}{0.009}
 & \ms{0.439}{0.010} & \ms{0.379}{0.027}
 & \ms{0.897}{0.005} & \ms{0.594}{0.008} & \ms{0.591}{0.009}
 & \ms{0.510}{0.009} & \ms{0.678}{0.010} \\
Focal Loss \cite{lin2017focalloss}
 & \ms{\textbf{0.829}}{0.011} & \ms{0.629}{0.016} & \ms{0.611}{0.017}
 & \ms{0.466}{0.020} & \ms{0.485}{0.027}
 & \ms{0.897}{0.007} & \ms{0.575}{0.011} & \ms{0.580}{0.010}
 & \ms{0.512}{0.010} & \ms{0.720}{0.011} \\
CCE Loss \cite{kim2021imbalanced}
 & \ms{0.800}{0.010} & \ms{0.579}{0.009} & \ms{0.574}{0.007}
 & \ms{0.431}{0.010} & \ms{0.534}{0.015}
 & \ms{0.883}{0.008} & \ms{0.579}{0.010} & \ms{0.581}{0.011}
 & \ms{0.506}{0.011} & \ms{0.691}{0.016} \\
LDAM Loss \cite{cao2019learning}
 & \ms{0.800}{0.024} & \ms{0.632}{0.025} & \ms{0.629}{0.021}
 & \ms{0.470}{0.019} & \ms{0.354}{0.060}
 & \ms{0.883}{0.010} & \ms{0.581}{0.011} & \ms{0.585}{0.012}
 & \ms{0.509}{0.012} & \ms{0.685}{0.019} \\
\hline
CB Loss \cite{cui2019balanced}
 & \ms{0.803}{0.012} & \ms{0.632}{0.015} & \ms{0.644}{0.014}
 & \ms{0.462}{0.015} & \ms{0.306}{0.041}
 & \ms{0.887}{0.009} & \ms{0.581}{0.013} & \ms{0.590}{0.015}
 & \ms{0.511}{0.014} & \ms{0.682}{0.017} \\
PF Loss \cite{du2021parameter}
 & \ms{0.790}{0.023} & \ms{0.635}{0.034} & \ms{0.641}{0.026}
 & \ms{0.465}{0.027} & \ms{0.323}{0.040}
 & \ms{0.879}{0.008} & \ms{0.584}{0.010} & \ms{0.595}{0.010}
 & \ms{0.512}{0.011} & \ms{0.661}{0.018} \\
AWB Loss \cite{yue2022toward}
 & \ms{0.802}{0.009} & \ms{\textbf{0.649}}{0.022} & \ms{0.648}{0.031}
 & \ms{0.467}{0.031} & \ms{0.299}{0.056}
 & \ms{0.884}{0.014} & \ms{0.585}{0.014} & \ms{0.593}{0.014}
 & \ms{0.512}{0.013} & \ms{0.691}{0.017} \\
DCAL \cite{kim2025long}
 & \ms{0.802}{0.014} & \ms{0.622}{0.013} & \ms{0.612}{0.016}
 & \ms{0.441}{0.018} & \ms{0.388}{0.033}
 & \ms{0.889}{0.008} & \ms{0.578}{0.010} & \ms{0.584}{0.011}
 & \ms{0.510}{0.013} & \ms{0.701}{0.019} \\
DWCE Loss
 & \ms{0.812}{0.015} & \ms{0.648}{0.014} & \ms{0.638}{0.017}
 & \ms{0.466}{0.017} & \ms{0.350}{0.047}
 & \ms{0.882}{0.011} & \ms{0.588}{0.015} & \ms{0.591}{0.015}
 & \ms{0.508}{0.016} & \ms{0.680}{0.021} \\
\hline
DAL (Ours)
 & \ms{0.799}{0.017} & \ms{0.647}{0.016} & \ms{\textbf{0.676}}{0.010}
 & \ms{\textbf{0.492}}{0.015} & \ms{\textbf{0.238}}{0.017}
 & \ms{0.888}{0.007} & \ms{\textbf{0.605}}{0.011} & \ms{\textbf{0.613}}{0.013}
 & \ms{\textbf{0.531}}{0.012} & \ms{\textbf{0.641}}{0.021} \\

\end{tabular}

\end{center}

\Caption{Test set Performance Across Four Imbalanced Benchmarks}{The methods are separated into non- or implicitly-weighted losses, explicitly class-weighted losses, and Ours. We report mean over 5 runs, and standard deviations in red. Best results per dataset and metric are highlighted in \textbf{bold}.}
\label{tab:uncertainty}
\end{table}

\begin{figure*}[p]
{
\centering
\subfigure[CE]{\includegraphics[width=0.24\textwidth]{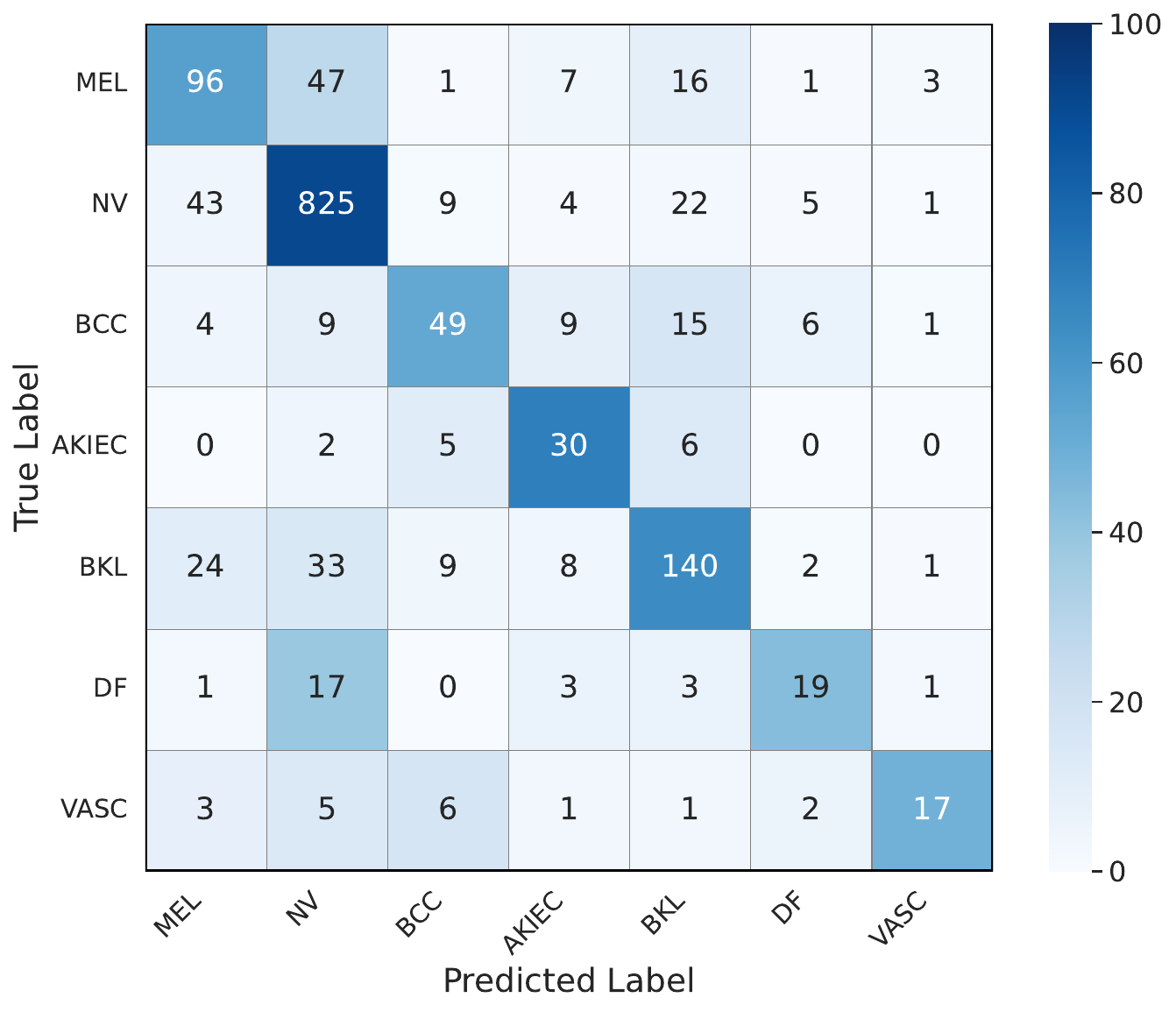}}\hfill
\subfigure[Focal]{\includegraphics[width=0.24\textwidth]{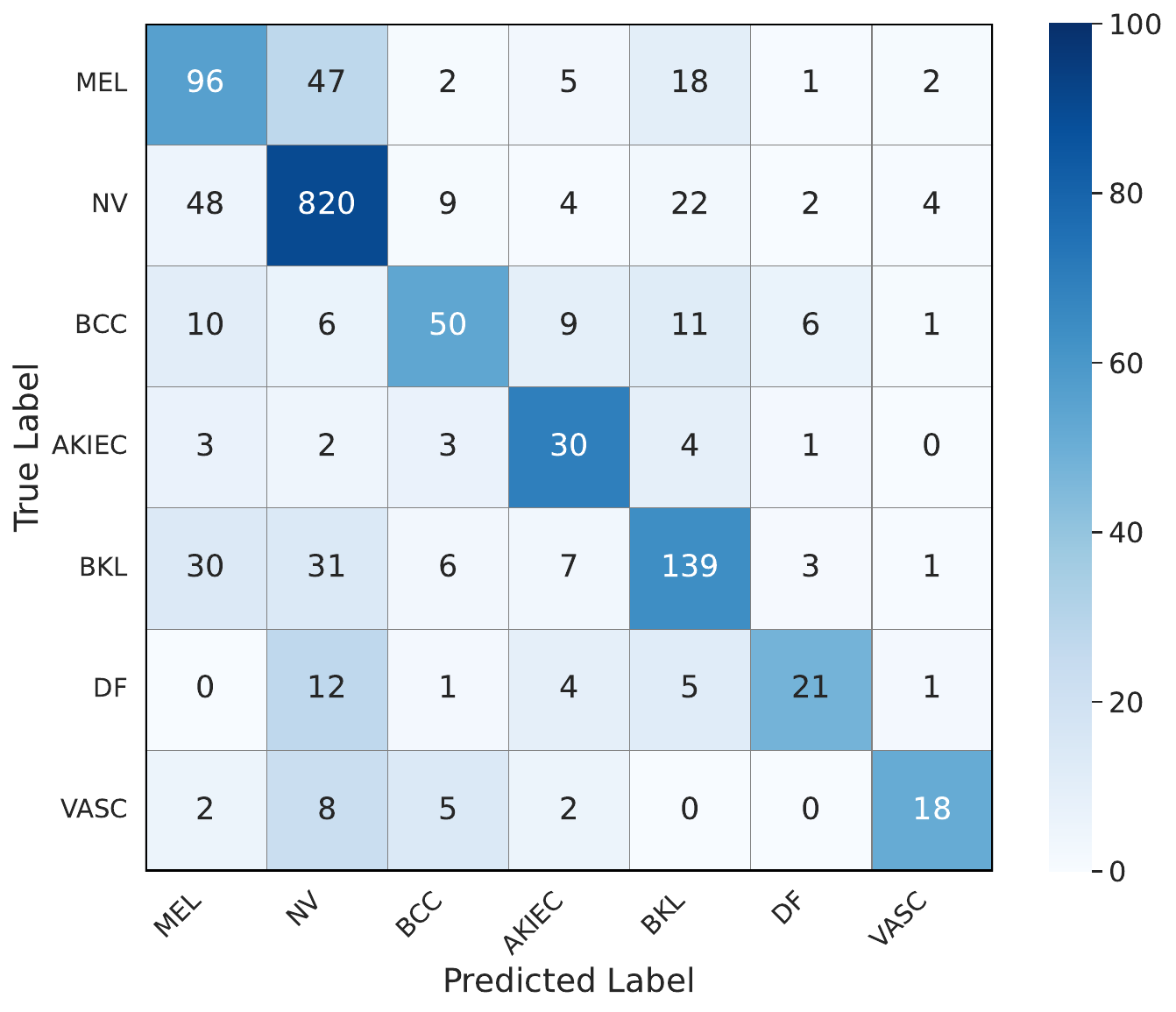}}\hfill
\subfigure[LDAM]{\includegraphics[width=0.24\textwidth]{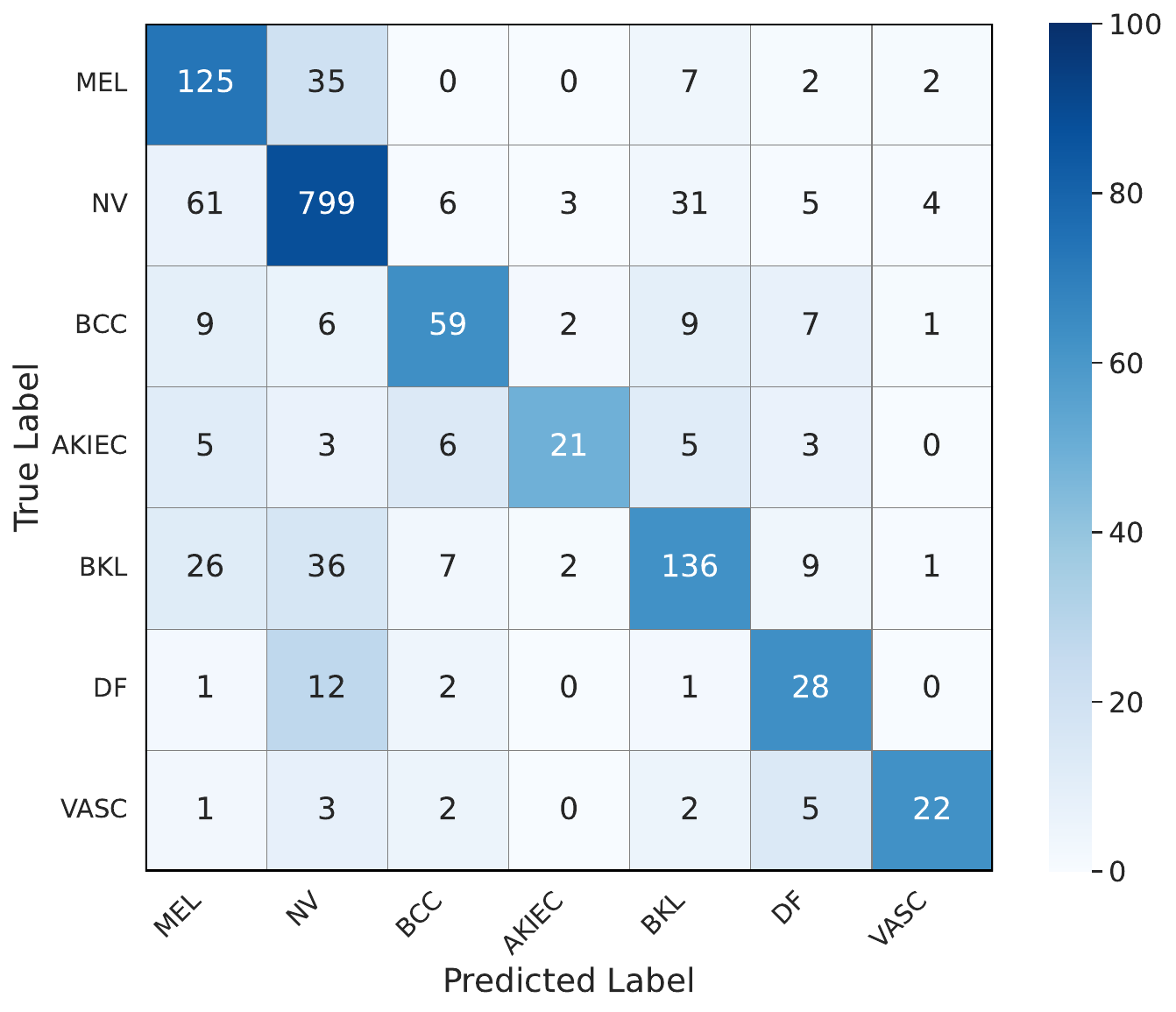}}\hfill
\subfigure[CB]{\includegraphics[width=0.24\textwidth]{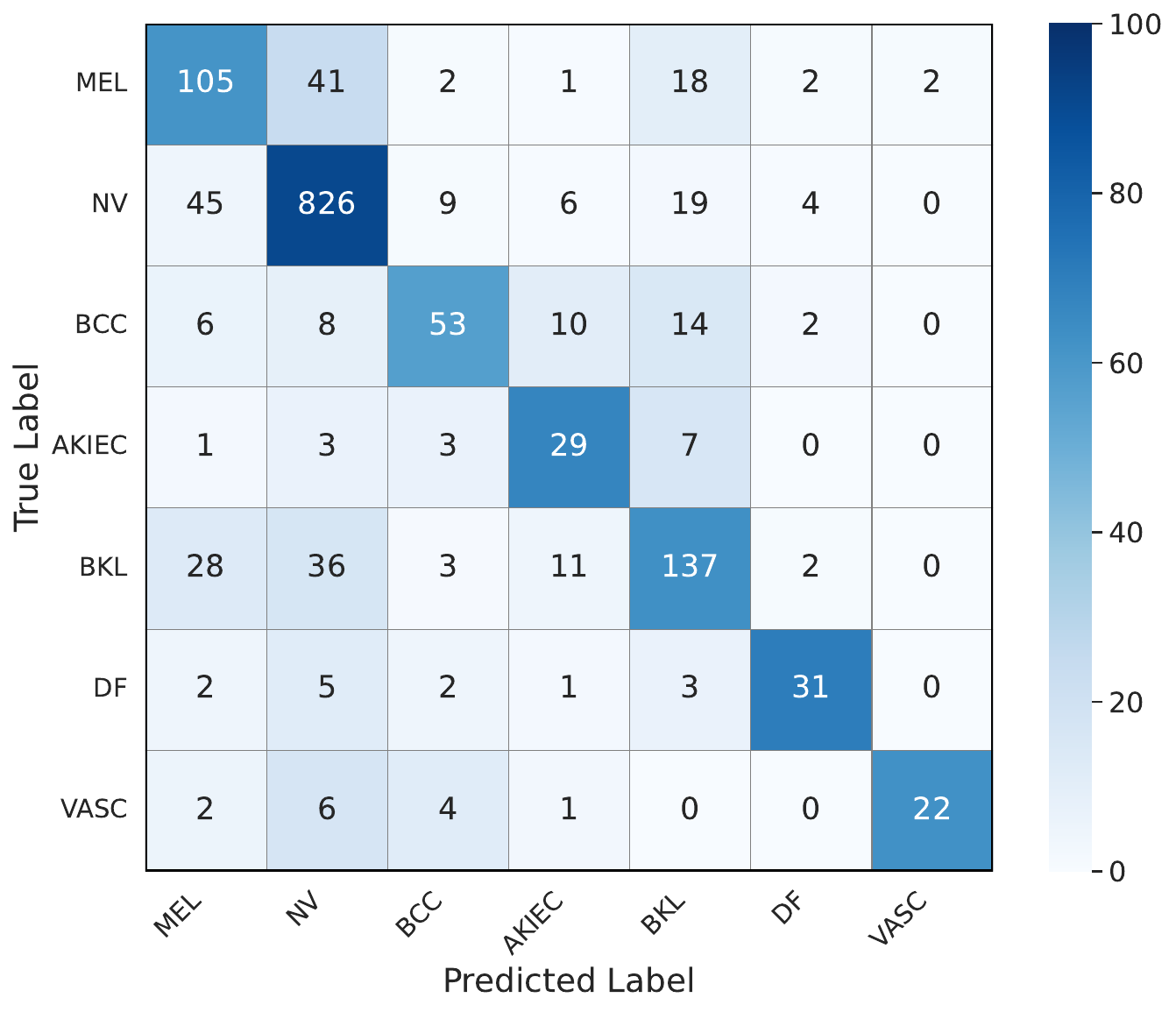}}\\[1ex]
\subfigure[CCE]{\includegraphics[width=0.24\textwidth]{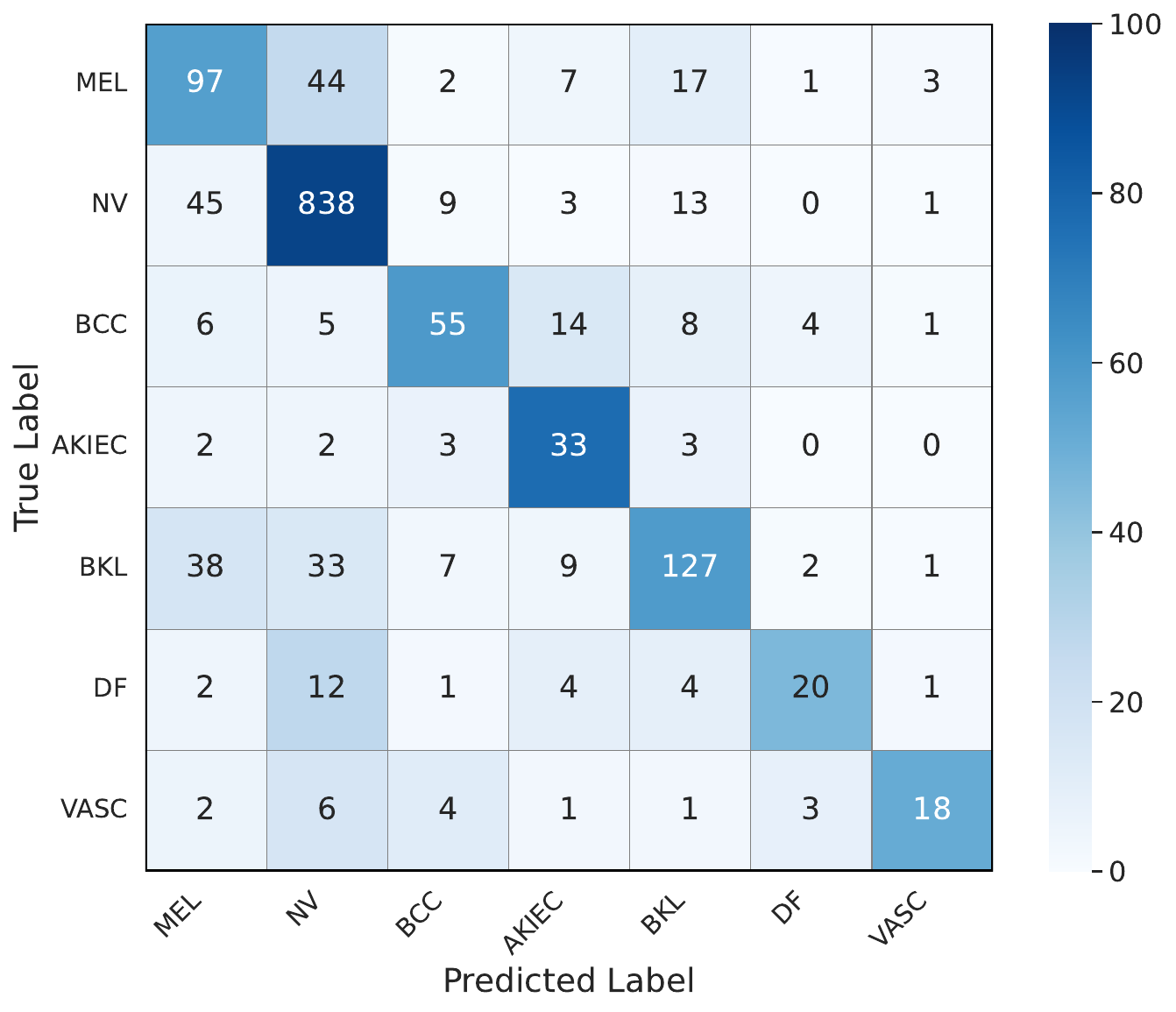}}\hfill
\subfigure[PF]{\includegraphics[width=0.24\textwidth]{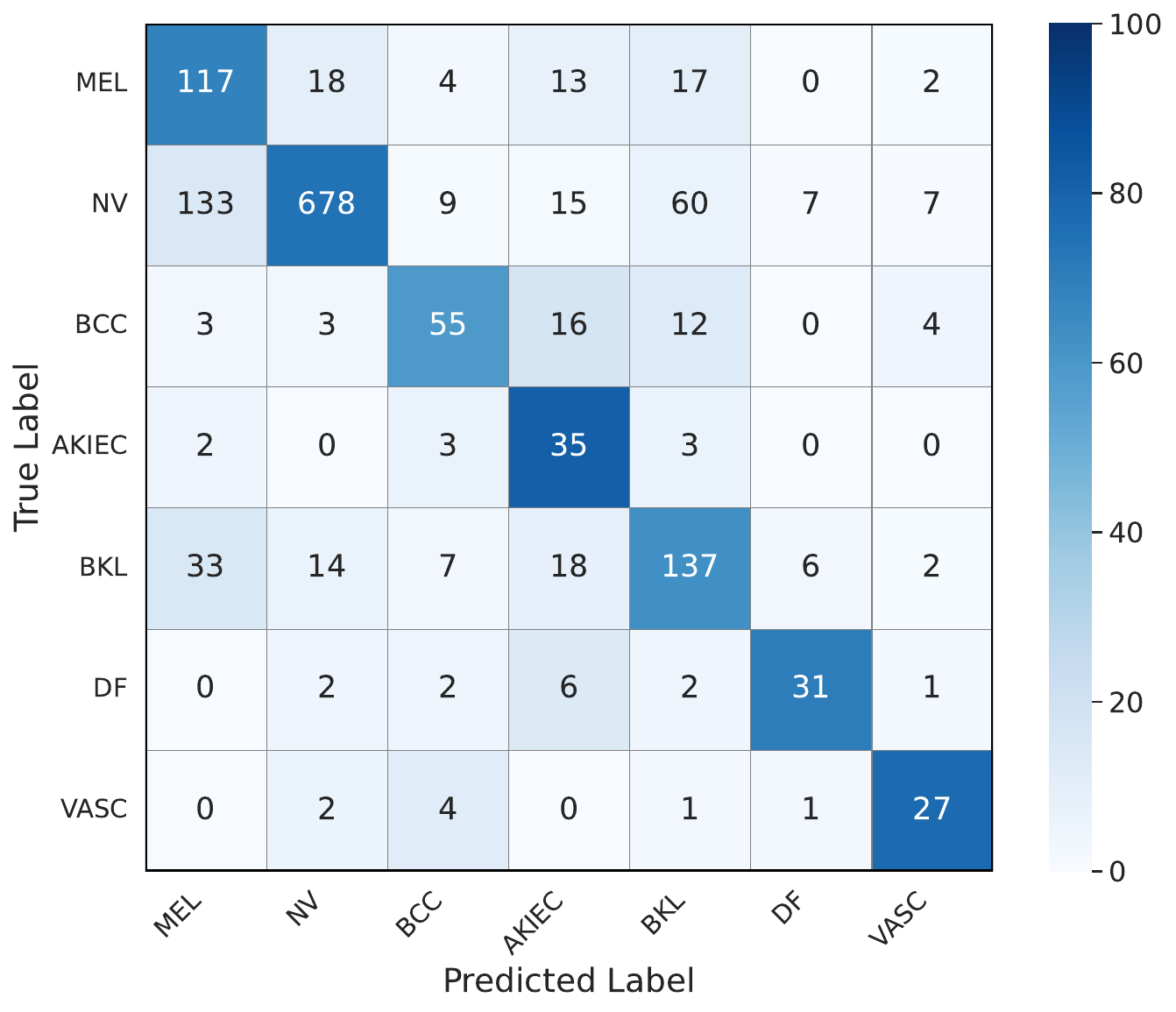}}\hfill
\subfigure[AWB]{\includegraphics[width=0.24\textwidth]{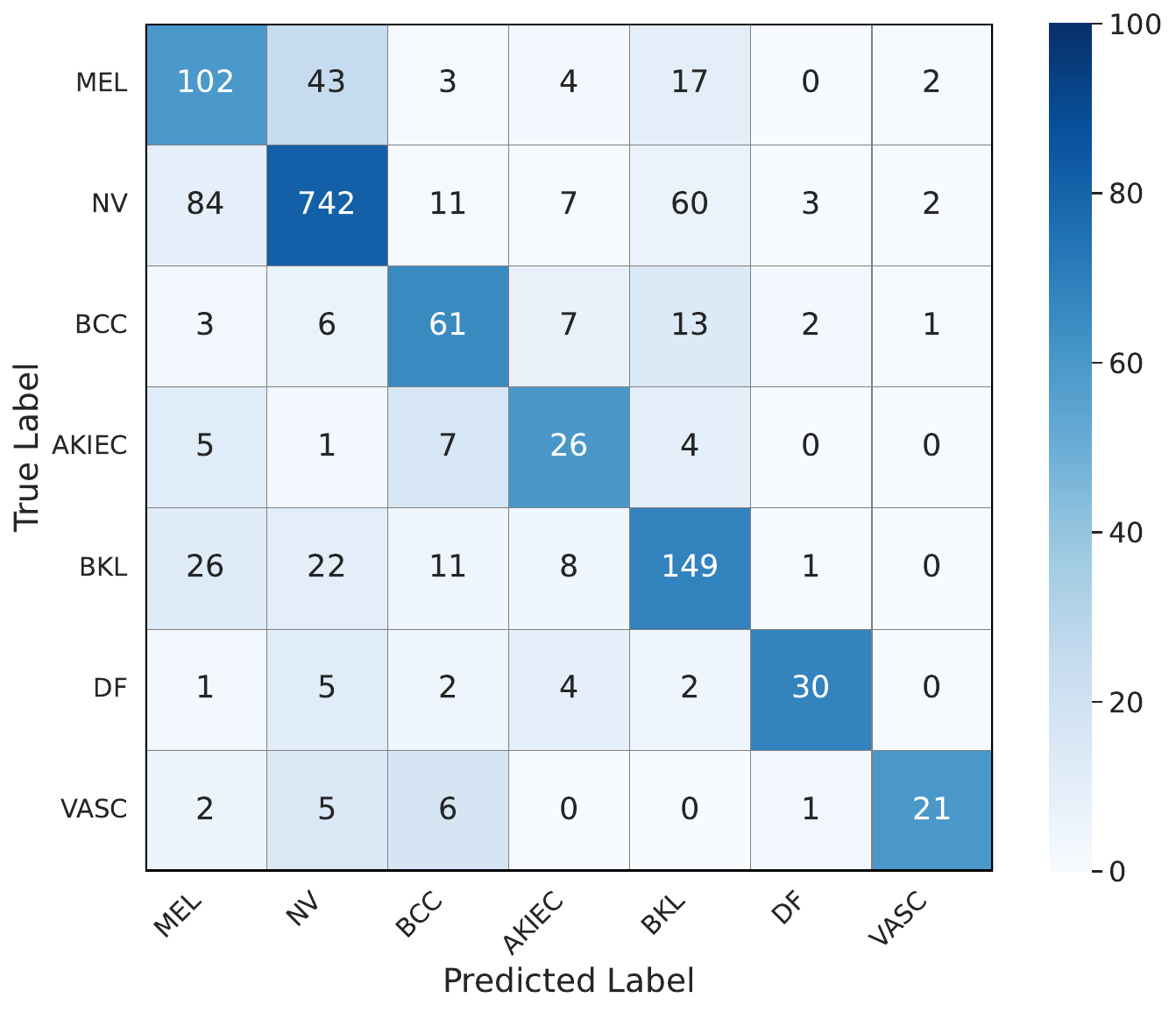}}\hfill
\subfigure[Ours]{\includegraphics[width=0.24\textwidth]{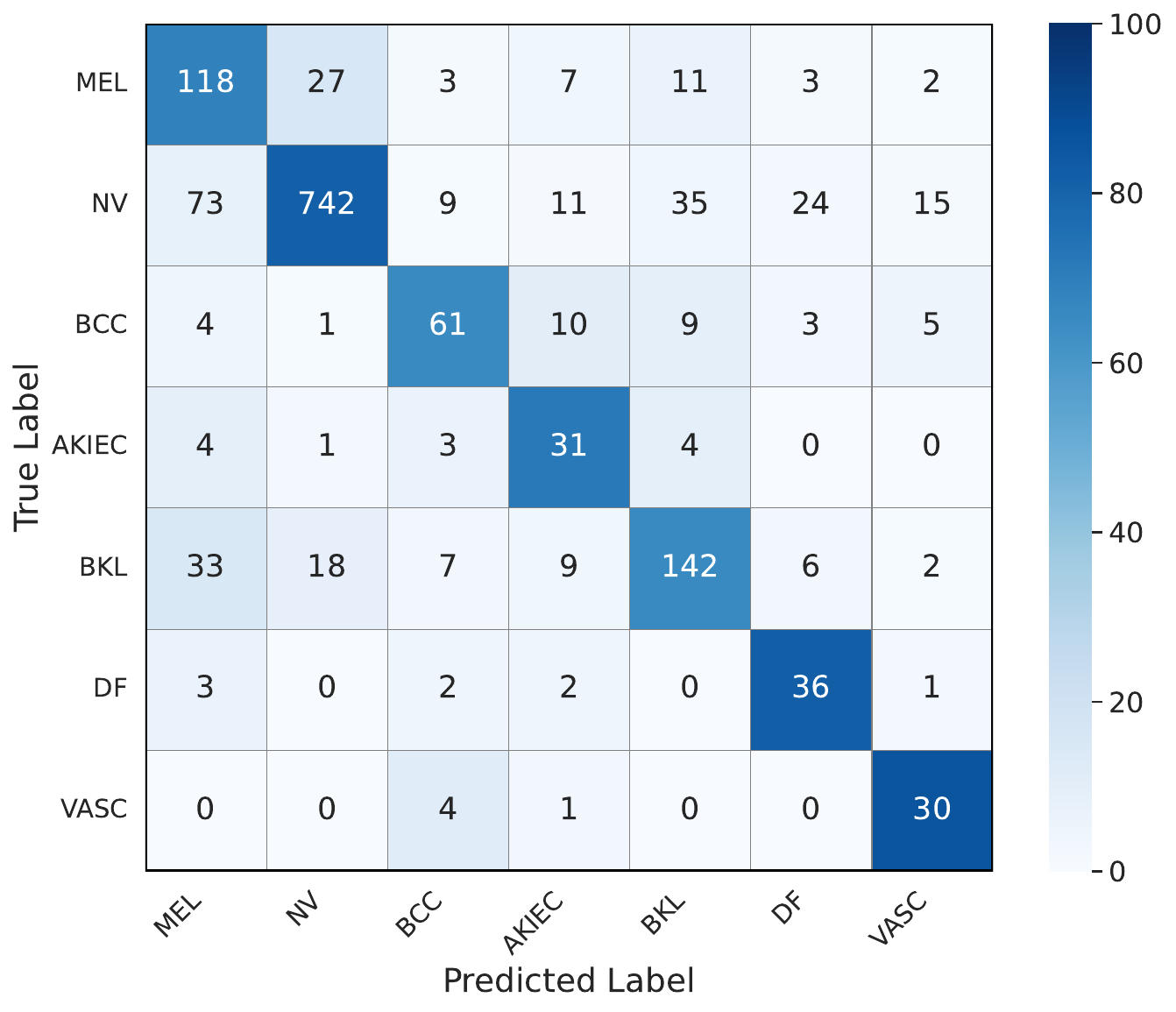}}
\par}
{\Caption[fig:isic2018_cm_all_losses]{Confusion Matrices for ISIC 2018}{We show complete confusion matrices of the seven classes in the ISIC~2018 test set for the different loss functions (best of five runs). The numbers show sample counts, while color saturation indicates relative performance.}}
\end{figure*}
\begin{figure*}[p]
{
\centering
\subfigure[CE]{\includegraphics[width=0.24\textwidth]{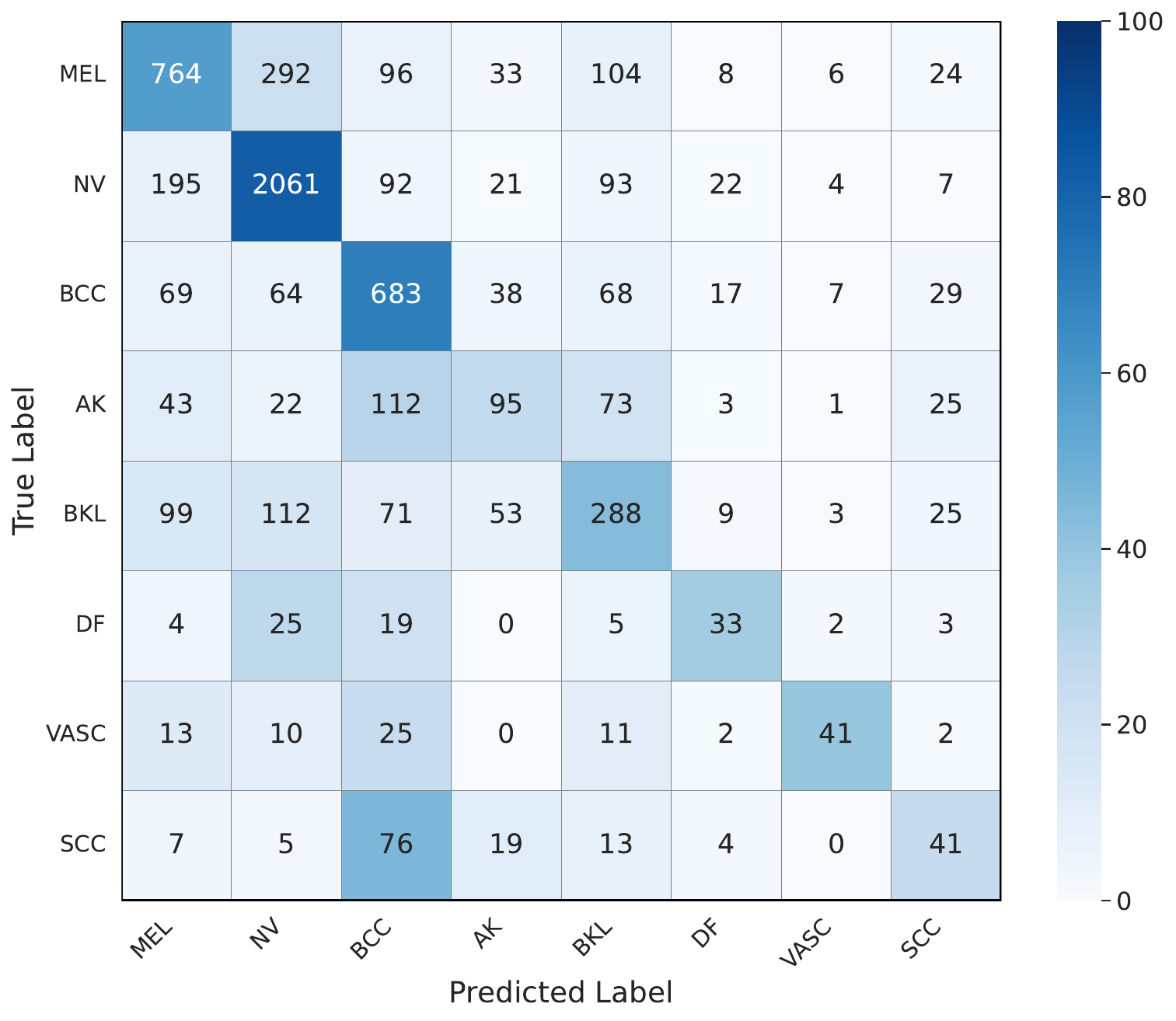}}\hfill
\subfigure[Focal]{\includegraphics[width=0.24\textwidth]{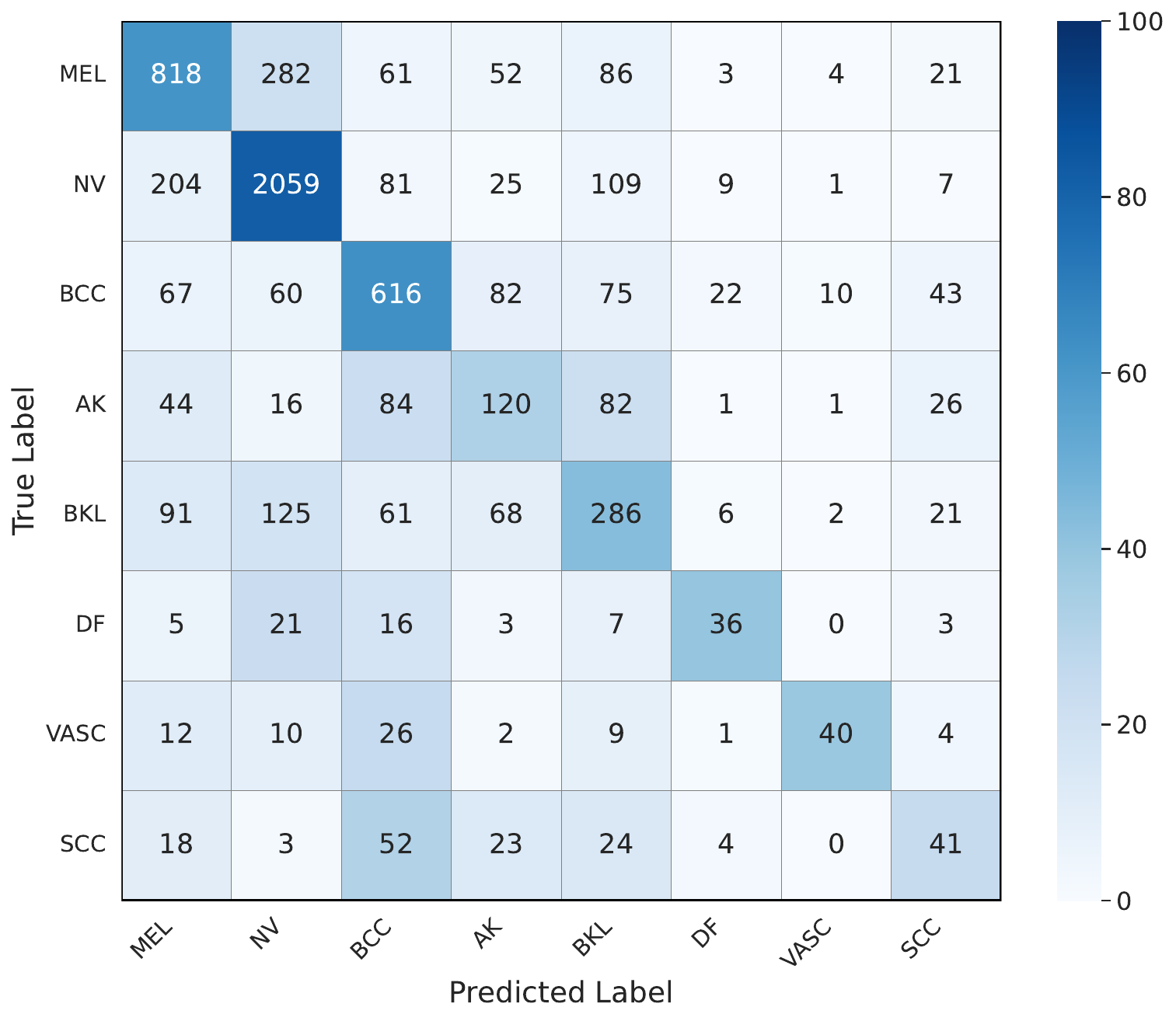}}\hfill
\subfigure[LDAM]{\includegraphics[width=0.24\textwidth]{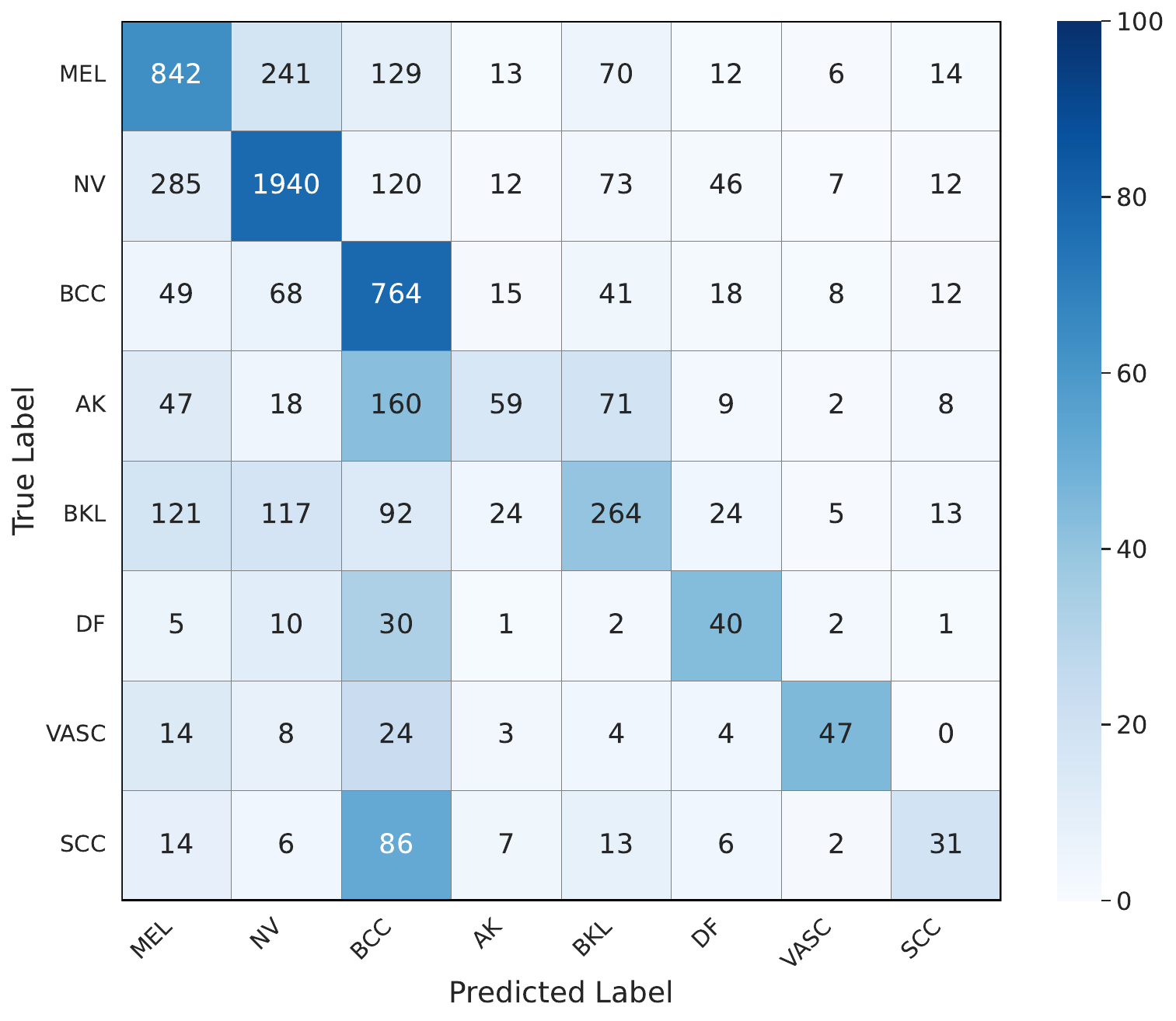}}\hfill
\subfigure[CB]{\includegraphics[width=0.24\textwidth]{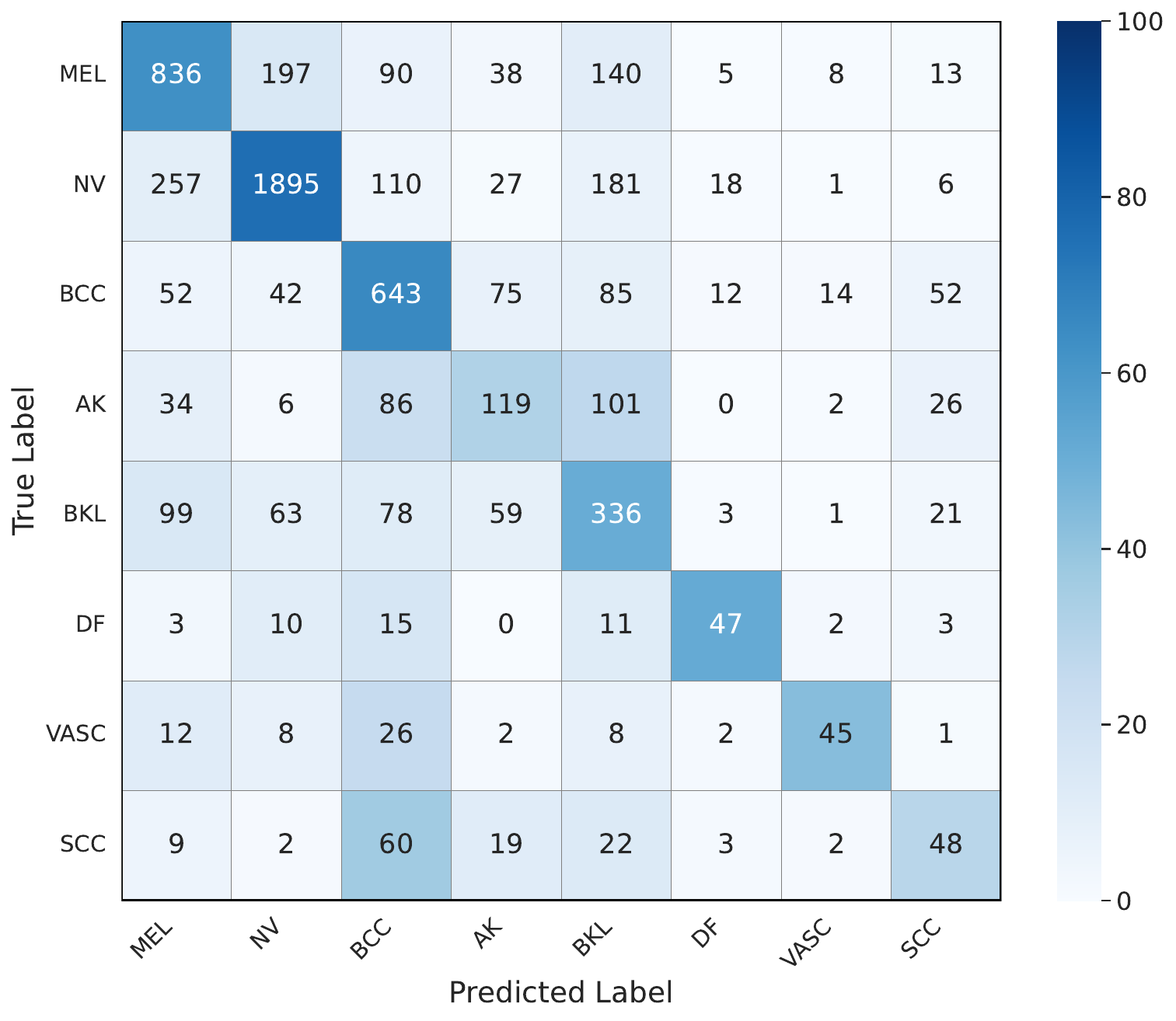}}\\[1ex]
\subfigure[CCE]{\includegraphics[width=0.24\textwidth]{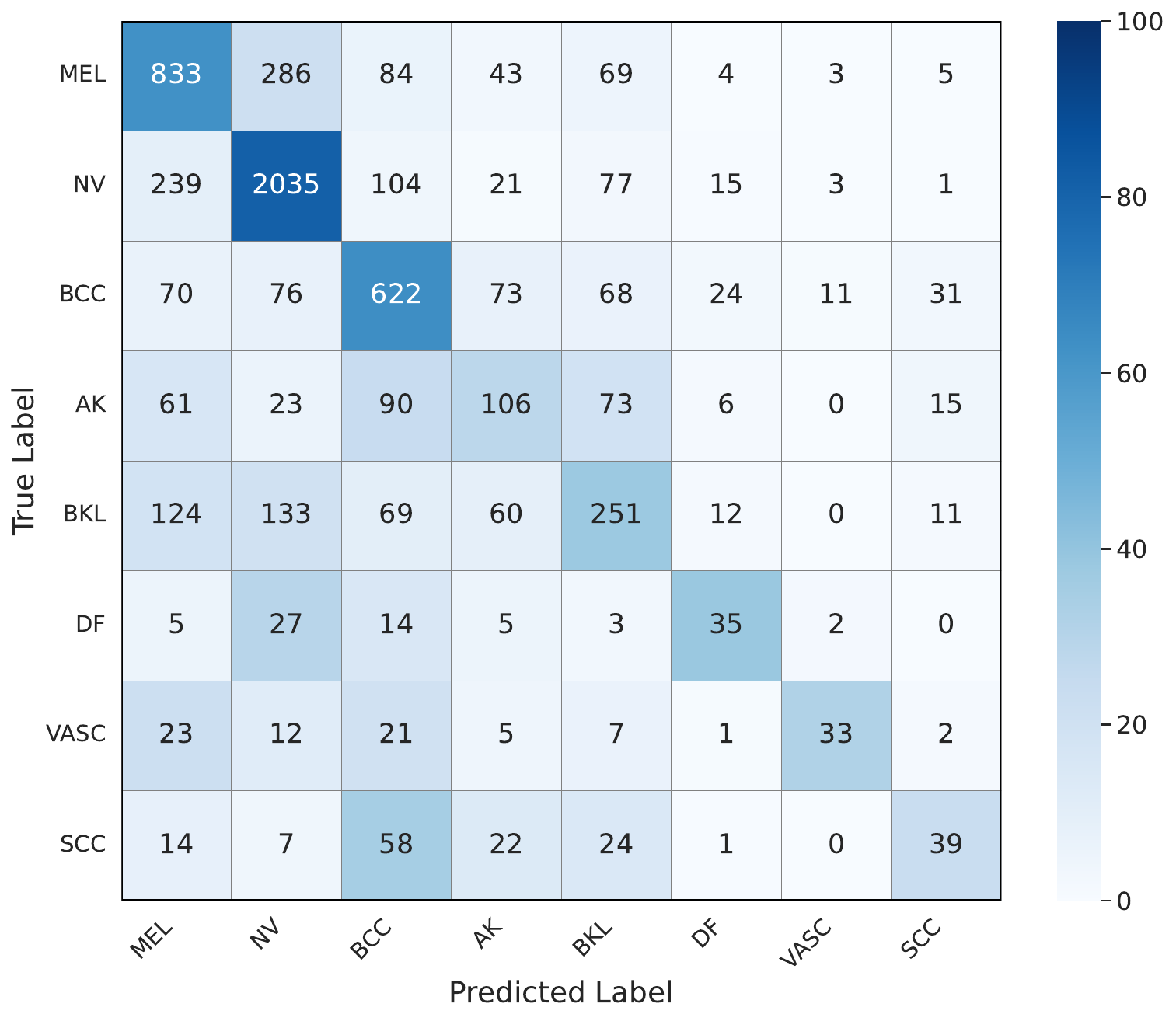}}\hfill
\subfigure[PF]{\includegraphics[width=0.24\textwidth]{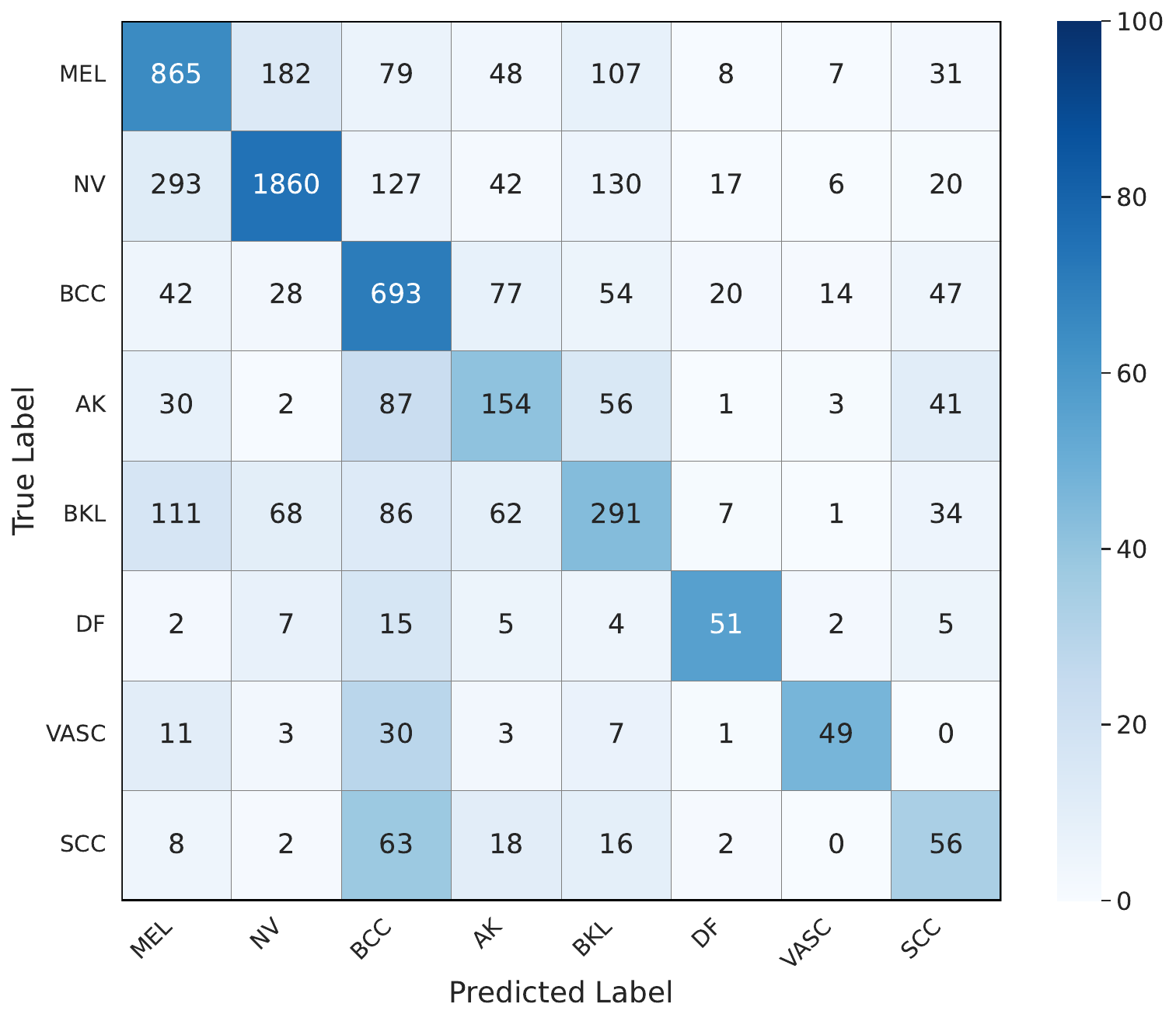}}\hfill
\subfigure[AWB]{\includegraphics[width=0.24\textwidth]{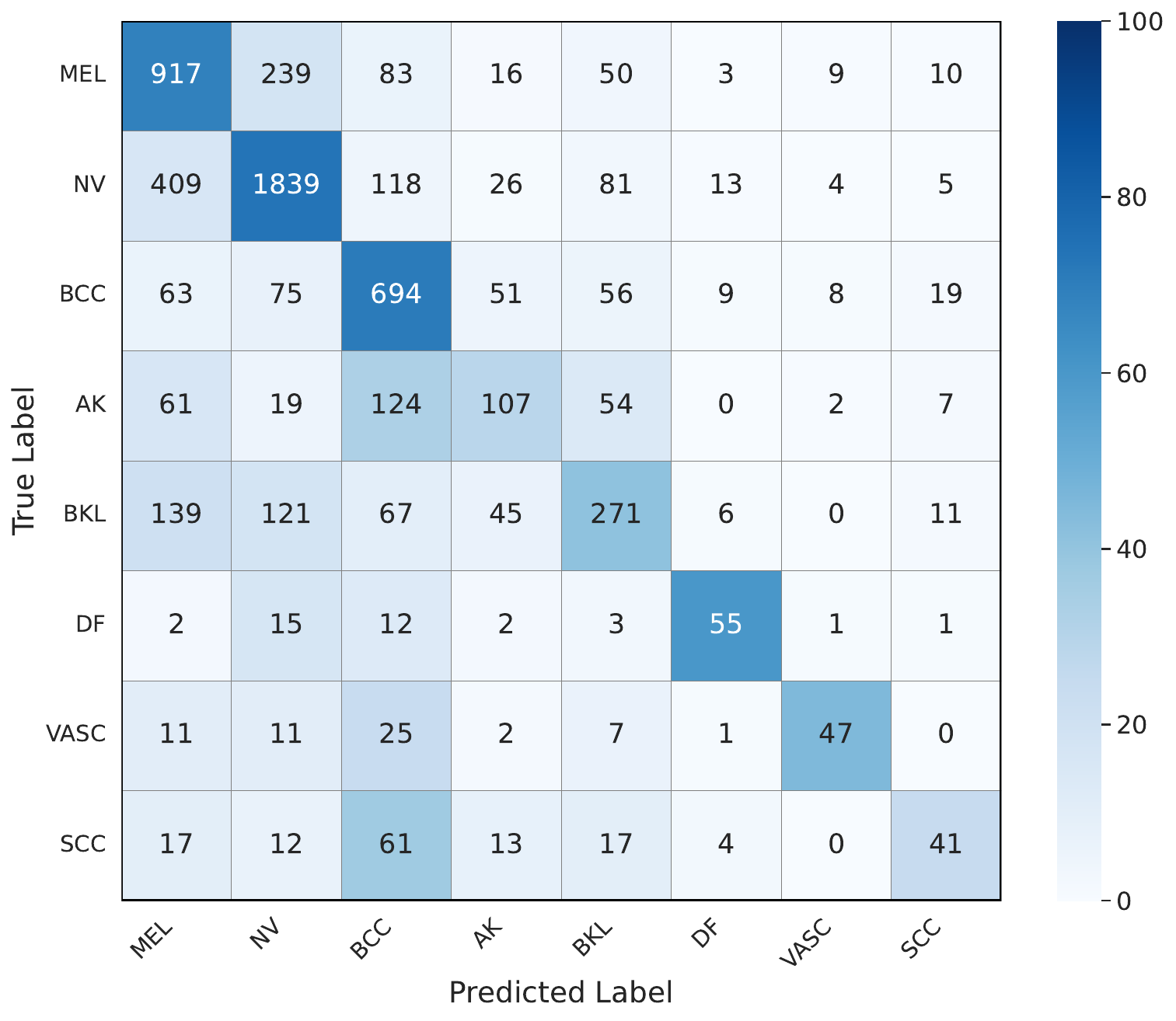}}\hfill
\subfigure[Ours]{\includegraphics[width=0.24\textwidth]{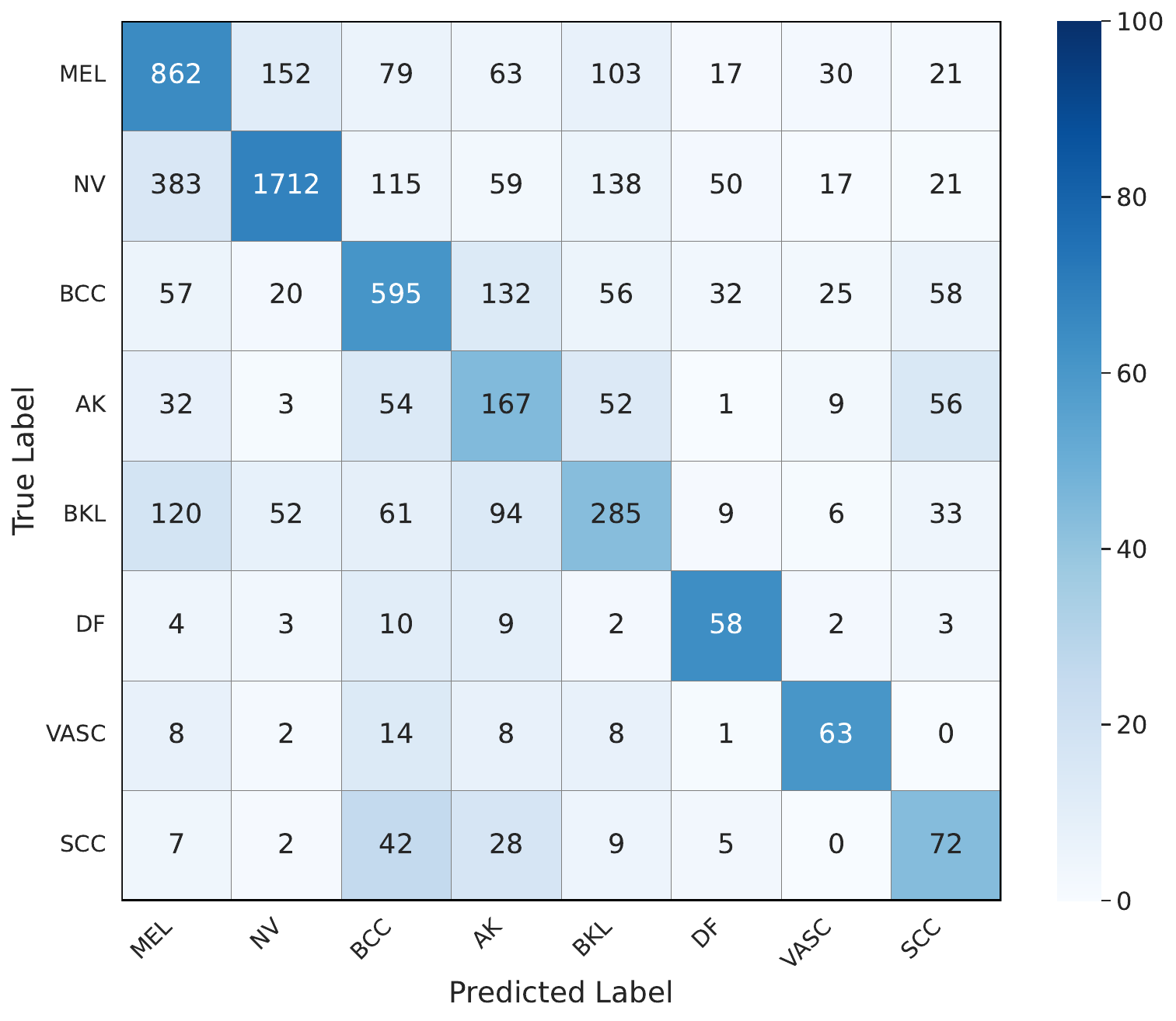}}
\par}
{\Caption[fig:isic2019_cm_all_losses]{Confusion Matrices for ISIC 2019}{We show the confusion matrices of the eight classes in the ISIC~2019 test set for the different loss functions (best of five runs). The numbers show sample counts, while color saturation indicates relative performance.}}
\end{figure*}

\section{Additional Results}

To provide more thorough insight into performance differences,
\figref{fig:isic2018_cm_all_losses} and \figref{fig:isic2019_cm_all_losses} show
the confusion matrices for all loss functions using the best-performing model
over five independent runs on the test sets of ISIC~2018 and ISIC~2019,
respectively. Clearly, the classes with fewer samples are better predicted than related loss functions, while performing slightly lower on majority classes.

We do not report further evaluation metrics for the APTOS dataset,
as it does not include a predefined test set and, therefore, test-set
generalization cannot be reliably assessed.
For Hyper-Kvasir, the number of classes is too large, which makes the comparison of confusion matrices difficult, so we do not include them here either.

\bibliography{sections/References,sections/Publications,sections/bibliography}